%% file: anonymous-submission-latex-2025.tex
\documentclass[letterpaper]{article} 
\usepackage{aaai25}  
\usepackage{times}  
\usepackage{helvet}  
\usepackage{courier}  
\usepackage[hyphens]{url}  
\usepackage{graphicx} 
\usepackage{natbib}  
\usepackage{caption} 
\usepackage{algorithm}
\usepackage{algorithmic}
\usepackage{amsmath}
\usepackage{multirow}
\usepackage{amssymb}
\usepackage{makecell}
\usepackage{graphicx}
\usepackage{newfloat}
\usepackage{listings}
\DeclareCaptionStyle{ruled}{labelfont=normalfont,labelsep=colon,strut=off} 
\floatstyle{ruled}
\newfloat{listing}{tb}{lst}{}
\floatname{listing}{Listing}
\title{Revisiting Tampered Scene Text Detection in the Era of Generative AI}
\author{
    Chenfan Qu\textsuperscript{\rm 1}, Yiwu Zhong\textsuperscript{\rm 2}, Fengjun Guo\textsuperscript{\rm 3, \rm 4}, Lianwen Jin\textsuperscript{\rm 1, \rm 4 \thanks{Corresponding author.}}\\
}
\affiliations{
    \textsuperscript{\rm 1}South China University of Technology\\
    \textsuperscript{\rm 2}The Chinese University of Hong Kong\\
    \textsuperscript{\rm 3}Intsig Information Co., Ltd\\
    \textsuperscript{\rm 4}INTSIG-SCUT Joint Lab on Document Analysis and Recognition\\

    202221012612@mail.scut.edu.cn, eelwjin@scut.edu.cn
}

\usepackage{bibentry}

\begin{document}

\maketitle

\begin{abstract}
The rapid advancements of generative AI have fueled the potential of generative text image editing, meanwhile escalating the threat of misinformation spreading. 
However, existing forensics methods struggle to detect unseen forgery types that they have not been trained on, underscoring the need for a model capable of generalized detection of tampered scene text. 
To tackle this, we propose a novel task: open-set tampered scene text detection, which evaluates forensics models on their ability to identify both seen and previously unseen forgery types. We have curated a comprehensive, high-quality dataset, featuring the texts tampered by eight text editing models, to thoroughly assess the open-set generalization capabilities. Further, we introduce a novel and effective pre-training paradigm that subtly alters the texture of selected texts within an image and trains the model to identify these regions. This approach not only mitigates the scarcity of high-quality training data but also enhances models' fine-grained perception and open-set generalization abilities. Additionally, we present DAF, a novel framework that improves open-set generalization by distinguishing between the features of authentic and tampered text, rather than focusing solely on the tampered text's features. Our extensive experiments validate the remarkable efficacy of our methods. For example, our zero-shot performance can even beat the previous state-of-the-art full-shot model by a large margin. Our dataset and code are available at https://github.com/qcf-568/OSTF.
\end{abstract}

\begin{figure}[t!]
  \centering
  \includegraphics[width=\linewidth]{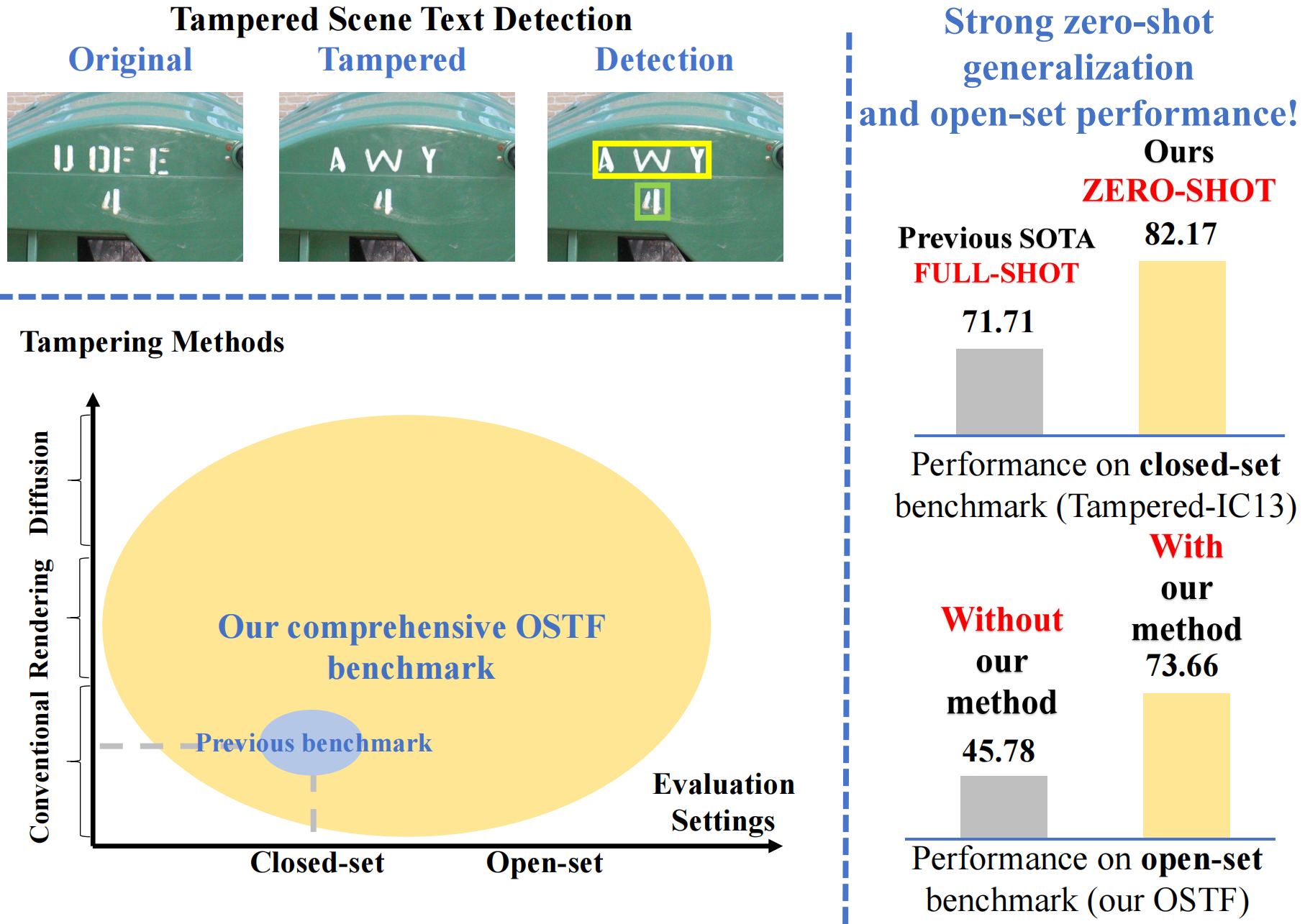}
  \caption{Tampered scene text detection aims to simultaneously detect real text (green box) and tampered text (yellow box) in the given image. In this paper, we introduce a novel task: open-set tampered scene text detection, where models are tested on both seen and unseen types of forgery. We also manually construct a comprehensive high-quality benchmark for this task. Moreover, we propose a simple-yet-effective method for this task, which shows strong zero-shot and open-set generalization ability.}
  \label{fig: teaser}
\end{figure}

\section{Introduction}

The rapid development of deep models sparks a generative AI revolution in computer vision, demonstrating remarkable progress in controllable editing~\cite{sun2023rethinking, qu2024towards}. However, the advancement of generative AI also leads to the spread of malicious fake information on text images, posing serious risks to social information  security~\cite{wang2022tic13, CVPR2023DocTamper}. Consequently, the detection of AI-tampered text has become a vital topic in recent years~\cite{qu2024omni}. It is crucial to develop effective methods for detecting AI-tampered text.

Recently, the Tampered-IC13 dataset~\cite{wang2022tic13} has been introduced to benchmark the detection methods for tampered scene text. Several promising methods have been proposed for this task, such as frequency domain feature extraction~\cite{wang2022tic13, CVPR2023DocTamper} and masked image modelling pre-training~\cite{peng2023viteraser, peng2023upocr}. Despite significant progress that has been achieved, the existing techniques are far from sufficient for real-world scenarios. We summarize the limitations as follows:

First, \textbf{the failure of the existing dataset} to reflect model performance in real-world scenarios.  The generative AI revolution has led to the continuous emergence of new text editing methods, producing increasingly realistic forgeries. The tampered texts in the Tampered-IC13 dataset were all forged using the outdated text editing model, SRNet~\cite{ACMMM_SRNet}, which is unlikely to be the real-world preference due to its relatively worse performance. Moreover, as the development of generative scene text editing is quite rapid, it is impractical to fit the forensics model to all types of the generative tampering methods. Therefore, the ability to generalize across unseen tampering methods and unseen scenarios becomes a crucial indicator of a model's practicality in real-world applications. Unfortunately, this open-set generalization ability cannot be adequately evaluated using the Tampered-IC13 benchmark.

Second, \textbf{the challenging nature of the tampered scene text detection task}. As introduced by pioneering work~\cite{wang2022tic13}, this task faces two main challenges. (1). The lack of high-quality training data. Existing text editing methods have a high probability of producing unsatisfactory outputs, requiring costly manual refinement in post-processing for visual consistency~\cite{wang2022tic13}. Training models on the purely generated samples without manual refinement will lead them to overfit to the obvious visual inconsistency, thereby making them difficult to generalize to real-world forgeries. (2). Enabling fine-grained perception. Tampered scene texts, after being generated by text editing models and further refined by human effort, become visually consistent with authentic texts~\cite{AAAI_MOSTEL, Anytext}. Only subtle texture anomalies may remain, which are challenging to detect. These two challenges have not been sufficiently addressed in previous works.

Third, \textbf{the poor open-set generalization of existing methods} for detecting tampered scene text. Forensic models oftentimes suffer from significant performance degradation on unseen types of forgeries. Such degradation is widely observed in other related fields, such as face anti-spoofing~\cite{FAS2}, deepfake detection~\cite{DFD1, DFD2}, image manipulation localization~\cite{sun2023rethinking}. Undoubtedly, no exception for the tampered scene text detection. However, none of the previous works pay attention to the open-set generalization of tampered scene text detection methods, resulting in a huge gap between the real-world applications.

To address these issues and to bridge the gap to real-world scenarios, we propose the following techniques:

1) To address the limitation of the existing benchmark, we manually construct a comprehensive high-quality benchmark for open-set tampered scene text detection, termed as Open-set Scene Text Forensics (OSTF). As shown in Figure~\ref{fig: teaser}, the tampered texts in our OSTF benchmark are tampered by a comprehensive set of text editing models, covering all the three types of scene text editing methods (conventional deep models, font rendering model, and diffusion models), successfully lining up with the recent development of generative AI. Except for the test setting of cross tampering methods, our OSTF benchmark also includes the test setting of cross source data, thus can better evaluate open-set generalization ability.

2) To simultaneously address the scarcity of high-quality training data, facilitate fine-grained perception, and improve open-set generalization, we propose a novel pre-training paradigm termed as Texture Jitter. In this paradigm, we subtly change the texture of the randomly selected text regions with elaborately designed operations to produce diverse types of texture anomalies, and train the model to localize these processed texts. Since the proposed Texture Jitter does not change the macro appearance, it can be applied directly to any type of image and will always produce perfectly visually consistent results. Therefore, the scarcity of high-quality training data can be significantly mitigated. In addition, since models trained with our Texture Jitter are forced to capture the subtle anomaly in the texture, the ability of fine-grained perception can be considerably improved. Moreover, 
by guiding the model to detect the texture anomaly rather than a specific type of tampering clue left by a particular tampering method, our Texture Jitter can also notably improve the model's open-set generalization. 

3) To further facilitate the models' ability of open-set generalization, we propose a novel pluggable framework, termed as Difference-Aware Forensics, inspired by the development of unsupervised anomaly detection~\cite{AD1, AD2}. The key idea is to learn a compact, robust representation for authentic text, and identify whether the input text is tampered or not by comparing its features with the learned authentic representation. By paying attention to the feature differences rather than only the features of the input text, models can better generalize to unseen forgeries.

We conduct extensive experiments on both the proposed OSTF benchmark and the widely-used Tampered-IC13~\cite{wang2022tic13} benchmark. Our method demonstrates strong generalization ability in these experiments. For example, the proposed method leads to a gain of 27.88 mean F-score on the open-set generalization ability in the OSTF benchmark. Moreover, the zero-shot version of our method even outperforms the full-shot version of the previous SOTA method UPOCR~\cite{peng2023upocr} by 10.46 mean IoU on the Tampered-IC13 benchmark.

In summary, the contributions of this paper are as follows:
\begin{itemize}
\item We propose a novel task, open-set tampered scene text detection, to meet the crucial demands. We manually construct a comprehensive high-quality benchmark for it.
\item We propose a novel pre-training paradigm for tampered scene text detection. It significantly mitigates the scarcity of high-quality training data, and notably improves capabilities for both fine-grained perception and open-set generalization.
\item We propose a novel pluggable framework that further improves open-set generalization on unseen forgery.
\item In-depth analysis and extensive experiments have verified the effectiveness of the proposed method.
\end{itemize}

\section{Related Works}
\textbf{Scene Text Editing}. Existing scene text editing methods can be divided into three types.~(1) Conventional deep models, which edit scene text in an E2E manner without utilizing diffusion models. SRNet~\cite{ACMMM_SRNet} is the first model to achieve E2E scene text editing. STEFANN~\cite{CVPR_STEFANN} proposed to edit scene text at char-level. MOSTEL~\cite{AAAI_MOSTEL} improved the visual quality of text editing with stroke-level masks and self-supervised learning on non-synthetic data. (2) Font-rendering based methods, which edit text with digital font files. DST~\cite{Shimoda_2021_ICCV} inpainted the original text region and rendered new text on it with the corresponding digital font file. This is very similar to the process of manually editing the target text using image processing software. (3) Diffusion methods, which leverage the power of diffusion models for realistic text tampering.  TextDiffuser~\cite{textdiffuser} generated text images with the given prompts, while DiffSTE~\cite{DiffSTE} manipulated the specific parts of the images with target texts. AnyText~\cite{Anytext} introduced Auxiliary Latent Module for higher visual quality. UDiffText~\cite{Udifftext} improved scene text editing with large-scale training data and text embedding. The rapid development of generative scene text editing techniques brings huge challenges and risks to social security~\cite{textdiffuser}. Therefore, it is essential to develop forensics models that can achieve open-set generalization on the text tampered by unseen methods.

\smallskip
\noindent \textbf{Tampered scene text detection} aims to localize tampered text on the given image. Wang. et al.~\cite{wang2022tic13} did the first work for tampered scene text detection, they proposed the first tampered scene text detection benchmark Tampered-IC13, they also proposed the S3R strategy and frequency domain modelling. However, this type of frequency domain modelling is likely to suffer from significant performance degradation on unseen types of forgery~\cite{AIGCD1}. Qu. et al.~\cite{CVPR2023DocTamper} introduced Selective Tampering Synthesis method and Document Tampering Detector model to improve tampered text detection in documents. However, these methods are not suitable for scene text due to the board variety in sizes and appearance. They also introduced the DocTamper synthetic dataset for documents, but it unable to benchmark model performance across unseen tampering methods and on AIGC-based tampering. Benefiting from more training data and more lenient evaluation metrics, Peng. et al.~\cite{peng2023upocr} achieved the highest mean F-score on the Tampered-IC13 benchmark by averaging real text detection scores. However, due to the lack of a specialized design, their performance for tampered text detection is still unsatisfactory. None of the existing work has explored open-set tampered scene text detection. To meet with the real-life requirements, we manually construct a comprehensive high-quality benchmark for open-set tampered scene text detection and propose novel, simple-yet-effective methods.

\section{OSTF Dataset and Benchmark}

\begin{table*}[ht!]
\centering
\caption{Comparison between our OSTF dataset and the previous dataset of Tampered Scene Text Detection.}
\setlength{\tabcolsep}{2pt}
\begin{tabular}{ccccccccccccccccccccc}
\hline
\multirow{2}{*}{Name} &  & \multirow{2}{*}{Year} &  & \multicolumn{3}{c}{Number of images} &  & \multicolumn{3}{c}{Number of texts} &  & \multicolumn{7}{c}{Tampering Methods} &  & \multirow{2}{*}{\begin{tabular}[c]{@{}c@{}} Cross source \\ Dataset \end{tabular}} \\ \cline{5-5} \cline{7-7} \cline{9-9} \cline{11-11} \cline{13-19}
 &  &  &  & All &  & Tampered &  & All &  & Tampered &  & Types &  & Conventional &  & Rendering &  & Diffusion &  &  \\ \cline{1-1} \cline{3-3} \cline{5-5} \cline{7-7} \cline{9-9} \cline{11-11} \cline{13-13} \cline{15-15} \cline{17-17} \cline{19-19} \cline{21-21} 
Tampered-IC13 &  & 2022 &  & 462 &  & 378 &  & 1944 &  & 995 &  & 1 &  & \checkmark &  & $\times$ &  & $\times$ &  & $\times$ \\
\textbf{Ours} & \textbf{} & \textbf{2024} & \textbf{} & \textbf{4418} & \textbf{} & \textbf{1980} & \textbf{} & \textbf{64858} & \textbf{} & \textbf{5018} & \textbf{} & \textbf{8} &  & \textbf{\checkmark} & \textbf{} & \textbf{\checkmark} & \textbf{} & \textbf{\checkmark} & \textbf{} & \textbf{\checkmark} \\ \hline
\end{tabular}
\label{datasetcomp}
\end{table*}

\begin{table*}[ht!]
\centering
\caption{The detailed statistics of the proposed OSTF benchmark.}
\setlength{\tabcolsep}{2pt}
\begin{tabular}{ccccccccccccccccccccc}
\hline
\multirow{3}{*}{\begin{tabular}[c]{@{}c@{}} Tampering \\ type \end{tabular}} &  & \multirow{3}{*}{\begin{tabular}[c]{@{}c@{}} Tampering \\ method \end{tabular}} &  & \multicolumn{7}{c}{Images} &  & \multicolumn{7}{c}{Text instances} &  & \multirow{3}{*}{Data source} \\ \cline{5-7} \cline{9-11} \cline{13-15} \cline{17-19}
 &  &  &  & \multicolumn{3}{c}{Authentic} &  & \multicolumn{3}{c}{Tampered} &  & \multicolumn{3}{c}{Authentic} &  & \multicolumn{3}{c}{Tampered} &  &  \\ \cline{5-5} \cline{7-7} \cline{9-9} \cline{11-11} \cline{13-13} \cline{15-15} \cline{17-17} \cline{19-19}
 &  &  &  & train &  & test &  & train &  & test &  & train &  & test &  & train &  & test &  &  \\ \cline{1-1} \cline{3-3} \cline{5-5} \cline{7-7} \cline{9-9} \cline{11-11} \cline{13-13} \cline{15-15} \cline{17-17} \cline{19-19} \cline{21-21} 
Rendering &  & DST~\cite{Shimoda_2021_ICCV} &  & 72 &  & 82 &  & 157 &  & 151 &  & 382 &  & 588 &  & 467 &  & 507 &  & ICDAR 2013 \\ \cline{1-1} \cline{3-3} \cline{5-5} \cline{7-7} \cline{9-9} \cline{11-11} \cline{13-13} \cline{15-15} \cline{17-17} \cline{19-19} \cline{21-21} 
\multirow{3}{*}{\makecell{Conventional\\deep models}} &  & SRNet~\cite{ACMMM_SRNet} &  & 29 &  & 55 &  & 200 &  & 178 &  & 342 &  & 607 &  & 507 &  & 488 &  & ICDAR 2013 \\
 &  & STEFANN~\cite{CVPR_STEFANN} &  & 182 &  & 181 &  & 47 &  & 52 &  & 721 &  & 946 &  & 128 &  & 149 &  & ICDAR 2013 \\
 &  & MOSTEL~\cite{AAAI_MOSTEL} &  & 168 &  & 172 &  & 61 &  & 61 &  & 628 &  & 882 &  & 221 &  & 213 &  & ICDAR 2013 \\ \cline{1-1} \cline{3-3} \cline{5-5} \cline{7-7} \cline{9-9} \cline{11-11} \cline{13-13} \cline{15-15} \cline{17-17} \cline{19-19} \cline{21-21} 
\multirow{4}{*}{\makecell{Diffusion\\models}} &  & DiffSTE~\cite{DiffSTE} &  & 174 &  & 181 &  & 55 &  & 52 &  & 683 &  & 943 &  & 166 &  & 152 &  & ICDAR 2013 \\
 &  & AnyText~\cite{Anytext} &  & 181 &  & 191 &  & 48 &  & 42 &  & 715 &  & 974 &  & 134 &  & 121 &  & ICDAR 2013 \\
 &  & UDiffText~\cite{Udifftext} &  & 129 &  & 132 &  & 100 &  & 101 &  & 471 &  & 772 &  & 378 &  & 323 &  & ICDAR 2013 \\
 &  & UDiffText~\cite{Udifftext} &  & 196 &  & 233 &  & 218 &  & 211 &  & 23737 &  & 22886 &  & 419 &  & 399 &  & TextOCR val \\
  &  & TextDiffuser~\cite{textdiffuserv2} &  & 40 &  & 40 &  & 123 &  & 123 &  & 2048 &  & 1515 &  & 123 &  & 123 &  & IC17, ReCTS val \\ \hline
\end{tabular}
\label{datasetdetail}
\end{table*}

\begin{figure*}[h!]
  \centering
  \includegraphics[height=120pt, width=\linewidth]{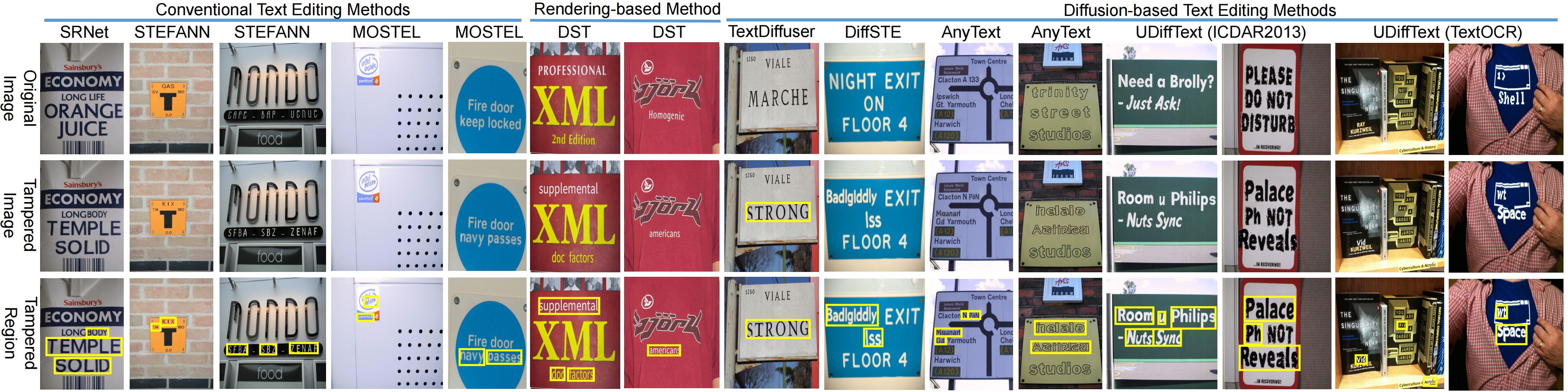}
  \caption{Samples in the proposed Open-set Scene Text Forensics (OSTF) dataset.}
  \label{fig: OTSE}
\end{figure*}

\textbf{Motivation}. In this era of generative AI, numerous new text editing models continuously emerge~\cite{textdiffuser, Udifftext}. However, the existing Tampered-IC13 benchmark only covers text tampered by the oldest text editing method SRNet, which can hardly be a real-world preference due to its relatively inferior performance. Moreover, the ability to detect the text tampered by unseen text editing model and that on unseen scenario is essential for forensics model in this era of generative AI. However, this ability totally cannot be evaluated in the Tampered-IC13 benchmark. To address the above issues, we manually construct a comprehensive high-quality new benchmark for tampered scene text detection, termed as OSFT, which includes text tampered by various latest text editing methods and cross source-dataset evaluation settings.

\subsection{Dataset Construction}
\noindent \textbf{Tampering Methods}. We take all the three types of generative tampering methods (conventional, font rendering, diffusion) into account, and select eight text editing methods, including SRNet~\cite{ACMMM_SRNet}, STEFANN~\cite{CVPR_STEFANN}, MOSTEL~\cite{AAAI_MOSTEL}, DST~\cite{Shimoda_2021_ICCV}, DiffSTE~\cite{DiffSTE}, AnyText~\cite{Anytext}, UDiffText~\cite{Udifftext}, Textdiffuser~\cite{textdiffuserv2}, as shown in Table~\ref{datasetdetail}. Given that the Tampered-IC13 dataset~\cite{wang2022tic13} already has reasonable forgeries tampered by SRNet, we take it as our `SRNet` part.

\noindent \textbf{Data Source}. We forge the text images from the ICDAR2013~\cite{karatzas2013icdar} with the selected eight text editing methods. To enable the cross-source dataset evaluation, we further edit the text images from the TextOCR~\cite{singh2021textocr} validation set with UDiffText, and the text images from the ICDAR2017~\cite{ic17mlt}, ReCTS~\cite{rects} validation sets with TextDiffuser.

\noindent \textbf{Manual Improvement}. To ensure the high quality of the tampered texts in our dataset, we edit all text instances using the chosen methods, and manually pick the most successful outputs. 
We further manually and elaborately improve the visual consistency of the picked texts using PhotoShop and GIMP. To ensure the high quality of the annotations, we manually update the bounding box for each tampered text to match its new appearance. We also check the labels via visualization to avoid errors.

\subsection{Dataset Statistics, Benchmark Settings, Highlights}
\noindent \textbf{Dataset Statistics}. As shown in Table~\ref{datasetcomp}, there are a total of 5018 tampered texts and 1980 tampered images in our OSTF dataset. The detailed statistics of it are shown in Table~\ref{datasetdetail}. 

\noindent \textbf{Evaluation Settings}. 
As shown in Table~\ref{datasetdetail}, there are 9 sessions in our dataset (ICDAR2013 tampered by 7 methods, TextOCR tampered by UDiffText, ICDAR2017 and ReCTS tampered by TextDiffuser). To evaluate both closed-set performance and open-set generalization, the models are \textbf{trained on one session of the training set} and \textbf{tested on all nine sessions of the testing set}. As a result, there are 9$\times$9=81 test settings, enabling three evaluation protocols: cross tampering methods, cross source dataset, and cross both tampering methods and source datasets. 

\noindent \textbf{Dataset Highlights}. The comparison between our OSTF dataset and previous dataset is shown in Table 1. Some of the samples in our dataset are shown in Figure~\ref{fig: OTSE}.
The main highlights of our dataset are as follows:

(1) \textbf{Comprehensive}. Our dataset includes all three types of generative text editing methods, keeping pace with the generative AI revolution. Our benchmark includes both the cross-AI-model and cross-dataset evaluation settings.

(2) \textbf{High Quality}. The tampered texts in our OSTF dataset are manually selected and elaborately enhanced for better visual consistency. The bounding boxes are manually adjusted to match the tampered texts.

(3) \textbf{Board Diversity}. The images in our OSTF dataset have various styles and resolutions. The texts in these images have various fonts and backgrounds, and are tampered by 8 different generative methods.

\section{Pre-training with Texture Jitter}

\textbf{Motivation}. Existing text editing methods often generate noticeable visual inconsistencies~\cite{wang2022tic13}, especially when dealing with complex fonts, unfamiliar languages, and diverse styles. Such failure to generate vivid tampered text leads to overfitting to obvious visual inconsistency. As the result, the trained models struggle to generalize their applicability to real-world scenarios where the visual appearance of the manipulated text is refined through human intervention and has minimized visual inconsistency. 

\begin{figure}[t!]
  \centering
  \setlength{\abovecaptionskip}{1pt}
  \setlength{\belowcaptionskip}{1pt}
  \includegraphics[height=110px, width=0.95\linewidth]{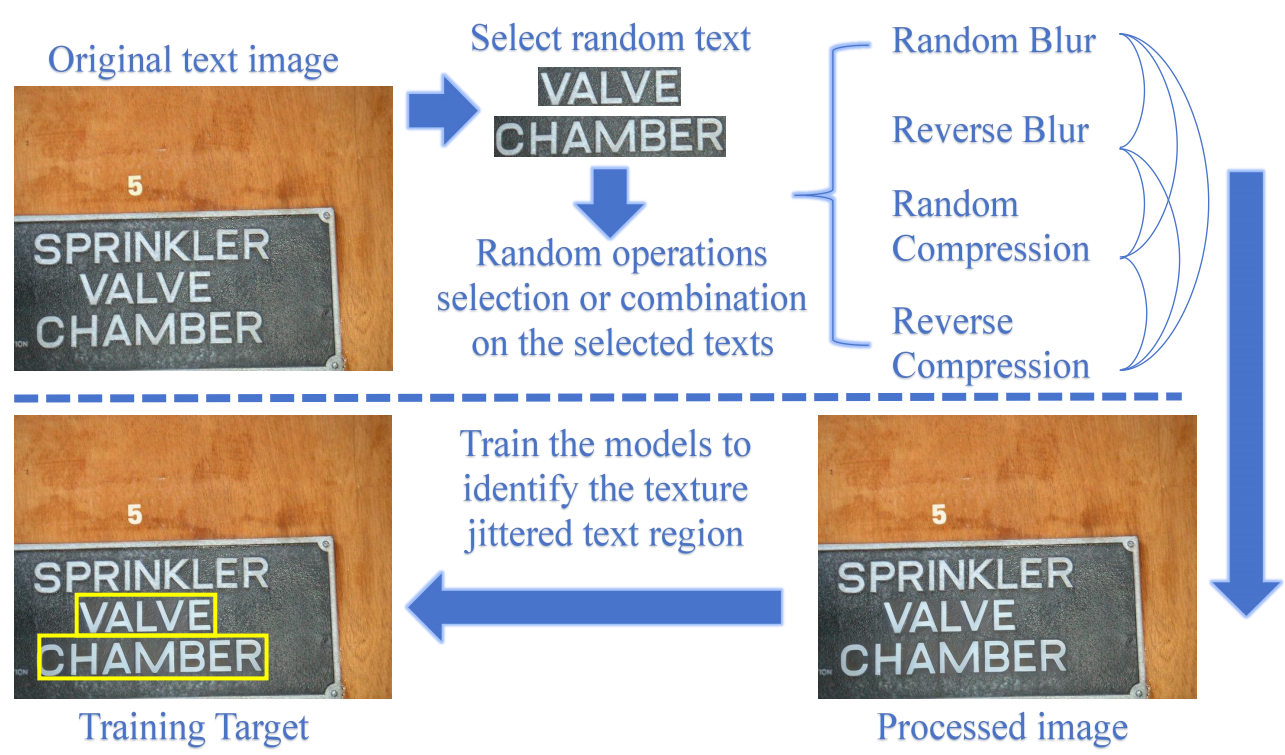}
  \caption{The pipeline of the proposed Texture Jitter.}
  \label{fig: TextureJitter}
\end{figure}

\smallskip
\noindent \textbf{Method}. Based on the above observations, we propose a \textbf{simple-yet-effective} method called Textual Jitter, as shown in Figure~\ref{fig: TextureJitter}. It slightly changes the texture of randomly picked texts while keeping their macro appearance unchanged, that is, the processed texts are almost the same as the original ones. Once this data processing is done, we train the models to localize the text regions and to identify whether their texture has been changed or not. Without changing the macro appearance, this approach can be applied directly to any text image and can always produce high-quality training data that is visually consistent and similar to the elaborately tampered text. Despite the simplicity, the proposed Texture Jitter can \textbf{simultaneously address three major issues} in tampered scene text detection:

(1) \textbf{The scarcity of high-quality training data} is significantly mitigated. Our Texture Jitter can automatically output tampered text that is visually consistent with the macro appearance. It can be directly applied to any type of image and any foreign language, and produce diverse and high-quality data, thereby addressing the data scarcity issue. 

(2) \textbf{The ability of fine-grained perception} is considerably improved. The text processed by our Texture Jitter is visually consistent with surrounding texts and its anomaly is obscure and difficult to capture. By training forensics models with our Texture Jitter, the models are encouraged to detect subtle anomalies and the ability of fine-grained perception can be improved.

(3) \textbf{The ability of open-set generalization} is notably enhanced. By training with our Texture Jitter, the models learn to identify the tampered text by judging whether the texture is abnormal, rather than relying on a specific feature of a tampering method, and thus achieve better open-set generalization on unseen forgery.

\smallskip
\noindent \textbf{Implementation}. The pipeline of our Texture Jitter is shown in Figure~\ref{fig: TextureJitter}. Given an input text image, we randomly select some text instances, and apply a random texture processing operation or the combination of multiple operations. The texture processing operations include random blur, reverse blur, random image compression, and reverse image compression. 
By doing this, various types of texture anomalies are created. The processed text instances are then spliced into the same positions as the original ones, with a smooth transition at the edges to avoid obvious visual anomalies. 
Further, to ensure a balance between the learning difficulty and a natural fusion between the original and processed regions, we also propose to adaptively adjust the intensity of jittering and the edge smoothing with rules, based on the size of the target text. With the proposed adaptive intensity, Texture Jitter can always produce satisfactory outputs with minimized visual inconsistency. More details are given in the appendix. The full implementation will be open-source. 

\section{Difference Aware Forensics}

\textbf{Motivation}. Forensics models often suffer from significant performance degradation on unseen forgeries~\cite{FAS2, sun2023rethinking}. Typically in the era of generative AI, advanced generative models are rapidly emerging and they can edit the texts in images in ways that the forensics models have never seen during training. Therefore, constructing a robust forensics model that can generalize across unseen forgeries is critical for real-life scenarios.

\begin{figure}[t!]
  \centering
  \setlength{\abovecaptionskip}{1pt}
  \setlength{\belowcaptionskip}{1pt}
  \includegraphics[height=110px, width=0.95\linewidth]{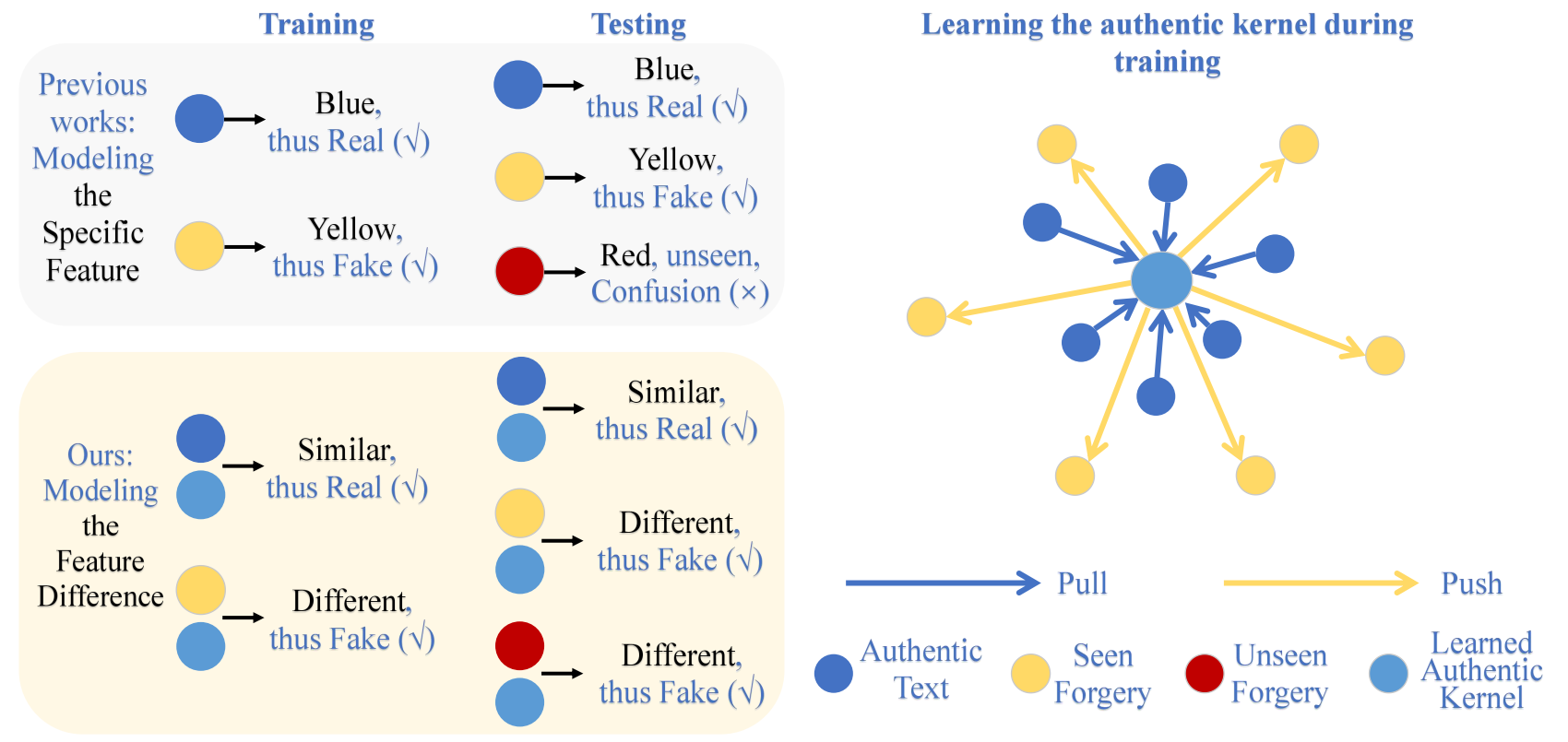}
  \caption{The key idea of our DAF is to model the feature difference rather than the input features themselves.}
  \label{fig: DAF idea}
\end{figure}

\smallskip
\noindent \textbf{Analysis}. The performance degradation on unseen forgeries is mostly caused by the training objective, a common binary classification task~\cite{FASAD}. As shown in the top left of Figure~\ref{fig: DAF idea}, during the training process, the models simply learn the specific features for the seen authentic class (blue circle) and the tampered class (yellow circle). When text is tampered by an unseen text editing method, the editing styles are never seen before and thus the features of the tampered text (red circle) are different from the seen ones. Consequently, the classifier is confused by new features of unseen forgeries, leading to poor performance.

\smallskip
\noindent \textbf{Key idea}. Although diverse text editing models will produce various tampering styles, the features extracted from the real text should remain similar. Moreover, the features of the tampered text can be regarded as anomalies since they are different from those of the authentic text. Inspired by this, we propose to model the difference between the input feature and the authentic feature for text forensics, instead of just relying on the individual input feature, as shown in Figure~\ref{fig: DAF idea}. Both the seen and unseen forgery features can be distinguished from the authentic feature, so that the confusion caused by the unseen forgery can be alleviated.

\begin{figure}[t!]
  \centering
  \setlength{\abovecaptionskip}{1pt}
  \setlength{\belowcaptionskip}{1pt}
  \includegraphics[height=110px, width=0.95\linewidth]{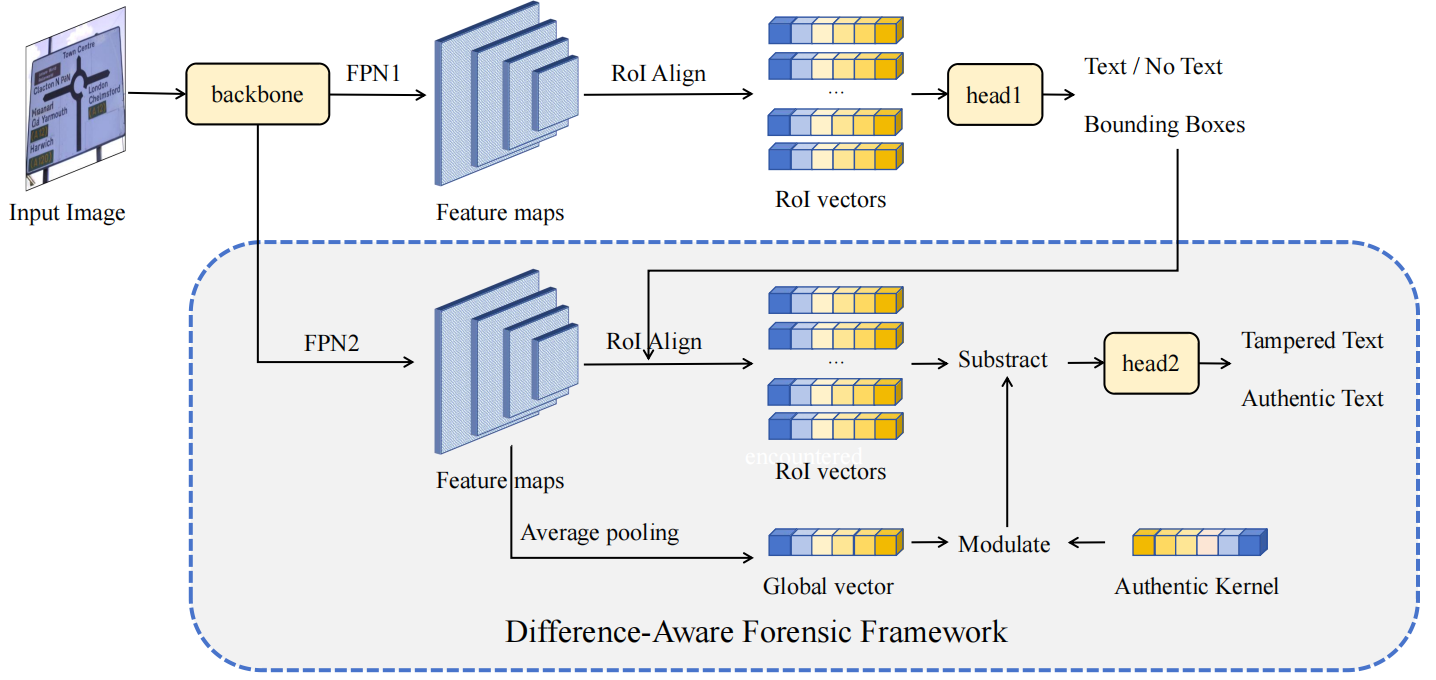}
  \caption{The proposed Difference-Aware Forensics (DAF). It builds on two-stage detectors such as Faster R-CNN.}
  \label{fig: DAF framework}
\end{figure}

\smallskip
\noindent \textbf{Implementation}. Specifically, we propose a novel framework DAF, as shown in Figure~\ref{fig: DAF framework}. Our DAF builds on the widely-used detection models (e.g., Faster R-CNN~\cite{ren2015fasterrcnn}) \textbf{by simply adding a forensics branch}, and is trained and tested in an E2E manner. The forensics branch shares the same backbone model as the original detection head, and extracts features for real/fake classification with an extra FPN network~\cite{lin2017feature}. This follows the S3R~\cite{wang2022tic13}, which performs text localization and forensics with separate networks. In the forensics branch, an authentic kernel is learned by pulling together the features of authentic text and pushing away the features of tampered text, which is supervised by L2 loss during training. More details are given in the appendix. \textbf{The intuition behind our design} is that the authentic kernel is approximately equivalent to the average of the features extracted from the real texts in the training set, and it can be regarded as the common representation for the real texts. For each input image, the learned authentic kernel is modulated by the global representation of the image with a linear layer to adapt to the image style. We subtract between the modulated authentic kernel and the RoI feature vectors. Finally, the subtracted features are fed into the classifier for final prediction.

\begin{table*}[ht!] 
\setlength\tabcolsep{1pt}
\setlength{\abovecaptionskip}{1pt}
\begin{minipage}[t]{0.48\linewidth}
\centering
\caption{Comparison study on the Tampered-IC13 dataset, 'z.' represents zero-shot and 'f.' represents full-shot.}{
\begin{tabular}{cccccccccc}
\hline
 &  & \multicolumn{2}{c}{Real Text} &  & \multicolumn{2}{c}{Fake Text} &  & \multicolumn{2}{c}{Average} \\ \cline{3-4} \cline{6-7} \cline{9-10} 
\multirow{-2}{*}{Method} &  & IoU & F &  & IoU & F &  & mIoU & mF \\ \cline{1-1} \cline{3-4} \cline{6-7} \cline{9-10} 
{DeepLabV3+ (f.)} &  & {48.1} & {65.0} & {} & {72.2} & {83.7} & {} & {60.2} & {74.4} \\
{HRNetv2 (f.)} &  & {43.3} & {60.4} & {} & {73.1} & {84.5} & {} & {58.2} & {72.4} \\
{BEiT-UPer (f.)} &  & {57.1} & {72.7} & {} & {70.9} & {83.0} & {} & {64.0} & {77.8} \\
{SegFormer (f.)} &  & {53.2} & {69.5} & {} & {77.8} & {87.5} & {} & {65.5} & {79.0} \\
{Swin-UPer (f.)} &  & 61.8 & 76.4 &  & 77.3 & 87.2 &  & 69.6 & 81.8 \\
UPOCR (f.) &  & 71.8 & 83.6 &  & 71.6 & 83.5 &  & 71.7 & 83.5 \\ \cline{1-1} \cline{3-4} \cline{6-7} \cline{9-10} 
Ours+Faster R-CNN \textbf{(z.)}&  & 79.9 & 88.8 &  & 84.5 & 91.6 &  & 82.2 & 90.2 \\
Ours+Cascade R-CNN \textbf{(z.)}&  & 79.1 & 88.3 &  & 84.3 & 91.5 &  & 81.7 & 89.9 \\
Ours+Faster R-CNN (f.) &  & \textbf{82.9} & \textbf{90.6} &  & 89.5 & 94.4 &  & \textbf{86.2} & \textbf{92.5} \\
{Ours+Cascade R-CNN (f.)} &  & {79.4} & {88.5} & {} & {\textbf{90.0}} & {\textbf{94.7}} & {} & {84.7} & {91.6} \\ \hline
\end{tabular}}
\label{tab: ticsegcomp}
\end{minipage}
\medspace
\begin{minipage}[t]{0.5\linewidth}
\centering
\caption{Comparison study on the Tampered-IC13 dataset, `S3R` represents the method proposed in~\cite{wang2022tic13}. 'z.' represents zero-shot and 'f.' represents full-shot.}
\vspace{+0.15cm}
\scalebox{0.95}{
\begin{tabular}{ccccccccccc}
\hline
 &  & \multicolumn{3}{c}{Real Text} &  & \multicolumn{3}{c}{Fake Text} &  & Avg. \\ \cline{3-5} \cline{7-9} \cline{11-11} 
\multirow{-2}{*}{Method} &  & P & R & F &  & P & R & F &  & mF \\ \cline{1-1} \cline{3-5} \cline{7-9} \cline{11-11} 
{S3R+EAST (f.)} &  & 50.5 & {27.3} & 35.5 & {} & 70.2 & {70.0} & {69.9} & {} & {52.7} \\
{S3R+PSENet (f.)} &  & 61.6 & {41.9} & 49.9 & {} & 79.9 & {79.4} & {79.7} & {} & {64.8} \\
{S3R+ATRR (f.)} &  & 76.7 & {54.6} & 63.8 & {} & 84.6 & {90.6} & {87.5} & {} & {75.7} \\
{S3R+CounterNet (f.)} &  & 77.9 & {54.8} & 64.3 & {} & 86.7 & {91.5} & {89.0} & {} & {76.7} \\
{DTD (f.)} &  & 0 & 0 & 0 &  & 92.1 & 89.3 & 90.7 &  & 45.4 \\ \cline{1-1} \cline{3-5} \cline{7-9} \cline{11-11} 
Ours+Faster R-CNN \textbf{(z.)} &  & 71.4 & 78.4 & 74.7 &  & 86.64 & 82.38 & 84.45 &  & 79.6 \\
Ours+Cascade R-CNN \textbf{(z.)} &  & 75.6 & 76.1 & 75.9 &  & 88.44 & 81.56 & 84.86 &  & 80.4 \\
Ours+Faster R-CNN (f.) &  & 80.5 & 81.7 & 81.1 &  & 91.44 & 96.31 & 93.81 &  & 87.5 \\
{Ours+Cascade R-CNN (f.)} &  & \textbf{83.0} & {\textbf{82.5}} & \textbf{82.7} & {} & \textbf{92.37} & {\textbf{96.72}} & {\textbf{94.49}} & {} & {\textbf{88.6}} \\ \hline
\end{tabular}}
\label{tab: ticdetcomp}
\end{minipage}
\end{table*}

\begin{table*}[ht!]
\centering
\caption{Ablation study on the OSTF dataset. `mP`, `mR`, `mF` denote mean precision, mean recall and mean F1-score respectively. `SCL` denotes using single-center-loss~\cite{SGL} in training. `DAF` denotes our Difference-Aware Forensics.}
\setlength{\tabcolsep}{1.5pt}
\scalebox{0.95}{
\begin{tabular}{ccccccccccccccccccccccccccccc}
\hline
\multicolumn{15}{c}{Ablation settings} &  & \multicolumn{13}{c}{Evaluation Score} \\ \cline{1-15} \cline{17-21} \cline{23-27} \cline{29-29} 
\multirow{2}{*}{\makecell{Num\\ber}} &  & \multicolumn{5}{c}{Pre-training} &  & \multicolumn{3}{c}{Framework} &  & \multicolumn{3}{c}{Base detector} &  & \multicolumn{5}{c}{Real Text} &  & \multicolumn{5}{c}{Tampered Text} &  & Average \\ \cline{3-7} \cline{9-11} \cline{13-15} \cline{17-21} \cline{23-27} \cline{29-29} 
 &  & \makecell{COCO\\Detection} &  & \makecell{Text\\Detection} &  & \makecell{Texture\\Jitter (Ours)} &  & \makecell{SCL} &  & \makecell{DAF\\(Ours)} &  & \makecell{Faster\\R-CNN} &  & \makecell{Cascade\\R-CNN} &  & mP &  & mR &  & mF &  & mP &  & mR &  & mF &  & mF \\ \cline{1-1} \cline{3-3} \cline{5-5} \cline{7-7} \cline{9-9} \cline{11-11} \cline{13-13} \cline{15-15} \cline{17-17} \cline{19-19} \cline{21-21} \cline{23-23} \cline{25-25} \cline{27-27} \cline{29-29} 
(1) &  & \checkmark &  &  &  &  &  &  &  &  &  & \checkmark & \textbf{} &  &  & 60.10 &  & 65.16 &  & 59.92 & \textbf{} & 49.28 &  & 31.87 &  & 34.82 & \textbf{} & 47.37 \\
(2) &  & \checkmark &  &  &  &  &  &  &  & \checkmark &  &  & \textbf{} & \checkmark &  & 63.56 &  & 68.01 &  & 63.38 & \textbf{} & 51.59 &  & 34.82 &  & 37.66 & \textbf{} & 50.52 \\
(3) &  &  &  & \checkmark &  &  &  &  &  &  &  & \checkmark &  &  &  & 73.98 &  & 73.35 &  & 71.34 &  & 59.84 &  & 44.10 &  & 45.78 &  & 58.56 \\
(4) &  &  &  & \checkmark &  &  &  &  &  & \checkmark &  &  &  & \checkmark &  & 75.25 &  & 73.97 &  & 72.66 &  & 63.07 &  & 46.79 &  & 48.01 &  & 60.34 \\
(5) &  &  &  &  &  & \checkmark &  &  &  &  &  & \checkmark &  &  &  & 76.62 &  & 76.43 &  & 74.87 &  & 77.33 &  & 70.57 &  & 71.96 &  & 73.41 \\
(6) &  &  &  &  &  & \checkmark &  & \checkmark &  &  &  & \checkmark &  &  &  & 77.52 &  & 76.28 &  & 75.26 &  & 76.94 &  & 71.05 &  & 72.32 &  & 73.79 \\
(7) &  &  &  &  &  & \checkmark &  &  &  & \checkmark &  & \checkmark &  &  &  & 78.54 &  & \textbf{76.55} &  & 75.98 &  & 77.55 &  & 72.62 &  & 73.66 &  & 74.82 \\
(8) &  &  &  &  &  & \checkmark &  &  &  & \checkmark &  &  &  & \checkmark &  & \textbf{82.54} &  & 74.63 &  & \textbf{76.74} &  & \textbf{80.42} &  & \textbf{72.43} &  & \textbf{74.96} &  & \textbf{75.85} \\ \hline
\end{tabular}}
\label{tab: OSTFdetcomp}
\end{table*}

\begin{figure}[ht]
  \centering
  \setlength{\abovecaptionskip}{1pt}
  \setlength{\belowcaptionskip}{1pt}
  \includegraphics[width=\linewidth]{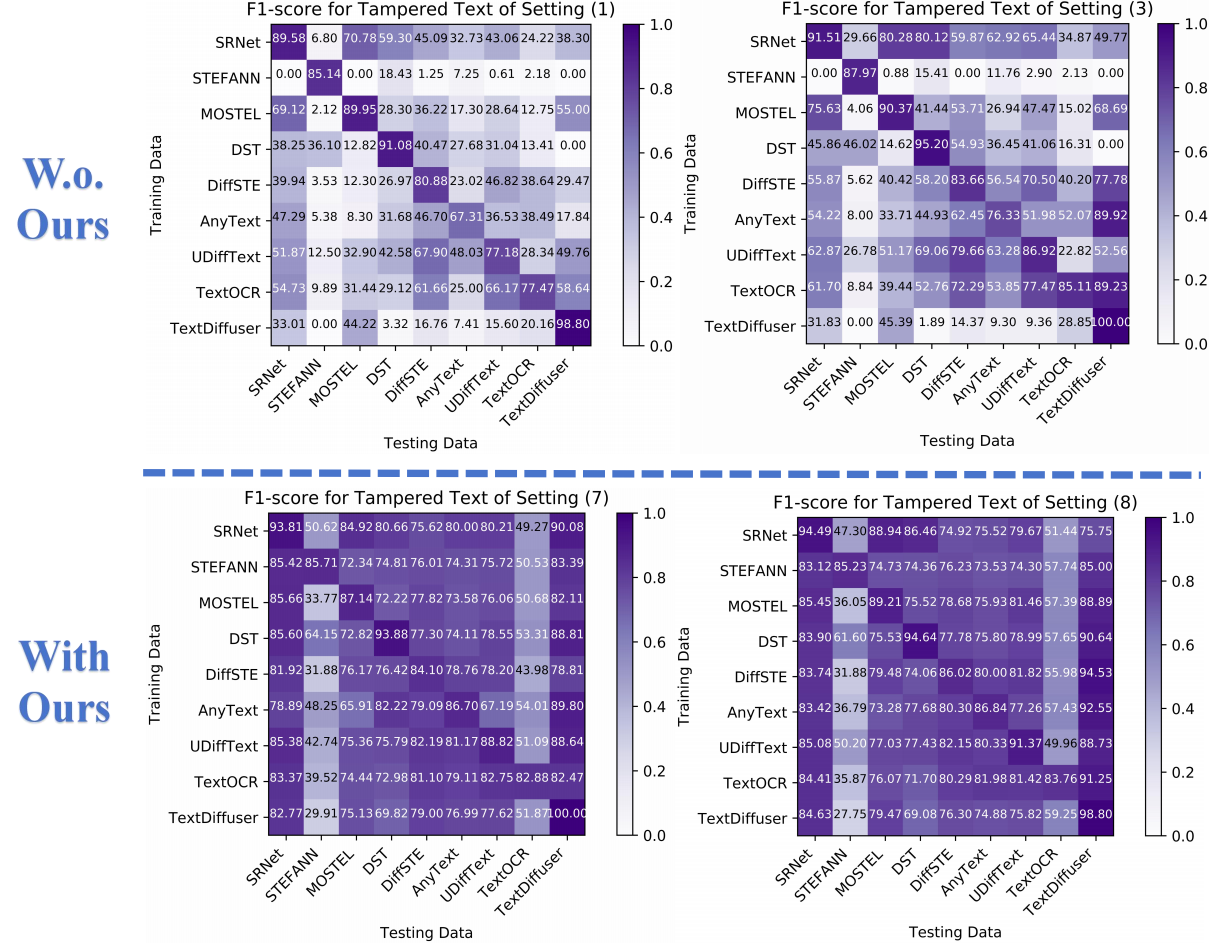}
  \caption{Ablation study on the OSFT dataset.}
  \label{fig: matrics}
\end{figure}

\section{Experiments}
\noindent \textbf{Implement Details}
We conduct experiments on both the well-acknowledged Tampered-IC13 benchmark~\cite{wang2022tic13} and our proposed OSTF benchmark. We first pre-train our model for 12 epochs with the proposed Texture Jitter on the same training sets as UPOCR~\cite{peng2023upocr}, including LSVT~\cite{lsvt}, ReCTS~\cite{rects}, ICDAR2013~\cite{karatzas2013icdar}, ICDAR2015~\cite{ic15}, ICDAR2017~\cite{ic17mlt}, TextOCR~\cite{singh2021textocr}, ArT~\cite{art}. The AdamW optimizer~\cite{adamw} with a learning rate initialized at 6e-5 and decaying to 1e-6 is used in the experiments. We then fine-tune the model using also the training sets of the Tampered-IC13~\cite{wang2022tic13} and the OSTF datasets respectively for 15k iterations with a batch size of 8. We adopt Swin-Transformer (Small)~\cite{liu2021swin} as the backbone following previous work~\cite{peng2023upocr}. The input image is resized to ensure that the shortest edge $\le$ 1024 and the longest edge $\le$ 1536.

\smallskip
\noindent \textbf{Evaluation Metric}.
For a fair comparison with the segmentation-based methods, we convert the bounding box predictions of our model into segmentation maps (detailed in the appendix) and calculate the Precision (P), Recall (R) and F1-score (F) at the pixel-level with the built-in functions of mmsegmentation~\cite{contributors2020mmsegmentation}  following the previous works~\cite{peng2023upocr}. For a fair comparison with the detection-based methods, we adopt P, R, F calculated at the instance-level with the official scripts provided by the Tampered-IC13~\cite{wang2022tic13}. We use these detection-based metrics for the ablation experiments.

\subsection{Comparison Study}
\noindent \textbf{Comparison with segmentation models.}  We compare the performance between our models and the SOTA segmentation-based forensic models, including DeepLabV3+~\cite{chen2018deeplabv3+}, HRNetV2~\cite{wang2020hrnetv2}, BEiT-UPer~\cite{bao2021beit}, SegFormer~\cite{xie2021segformer}, Swin-UPer~\cite{liu2021swin} and UPOCR~\cite{peng2023upocr}. The comparison results on the Tampered-IC13 benchmark~\cite{wang2022tic13} are shown in Table~\ref{tab: ticsegcomp}. The \textbf{zero-shot variants} of our models can even notably outperform the \textbf{full-shot version} of the previous SOTA model UPOCR by \textbf{more than 10 points} mIoU. The ``zero-shot'' represents our pre-trained model that has never seen any tampered text used in fine-tuning. Its strong zero-shot generalization ability is attributed to the effectiveness of our methods. Our full-shot models achieve higher performance.

\smallskip
\noindent \textbf{Comparison with detection models.} We compare the performance between our models and the SOTA detection-based forensic models, including S3R+EAST~\cite{zhou2017east}, S3R+PSENet~\cite{wang2019psenet}, S3R+ATRR~\cite{wang2019atrr}, S3R+CounterNet~\cite{wang2020contournet} and DTD~\cite{CVPR2023DocTamper}. As shown in Table~\ref{tab: ticdetcomp}, the zero-shot variants of our models also outperform the full-shot version of the state-of-the-art (SOTA) model by about 3 points mean F-score. The full-shot variants of our models further significantly outperform the state-of-the-art model (e.g. the full-shot version of `Ours+Cascade R-CNN` achieves 88.62 mean F-score, about \textbf{12 points higher} than the SOTA of 76.66), demonstrating strong generalization.

\subsection{Ablation Study} 
The ablation results on the OSFT dataset are shown in Table~\ref{tab: OSTFdetcomp} and Figure~\ref{fig: matrics}. In Table~\ref{tab: OSTFdetcomp}, the `mP`, `mR`, `mF` are calculated by averaging the P, R, F of all the 81 test settings in the OSTF dataset. In Figure~\ref{fig: matrics}, the four matrices are from the ablation settings (1), (3), (7), (8) in Table~\ref{tab: OSTFdetcomp}, respectively.

In Table~\ref{tab: OSTFdetcomp}, settings (1), (3), (5) and (2), (4), (8) are designed to verify the effectiveness of our Texture Jitter paradigm. The model pre-trained with our Texture Jitter (setting (5)) gets 71.96 mF-score on tampered text detection, 26.18 points higher than the baseline pre-trained with the same training configuration but only with the common text detection task (setting (3)), and even 37.14 points higher than the baseline initialized with the official COCO detection pretrained weights (setting (1)). Similarly, huge improvement on Cascade R-CNN~\cite{cai2018cascadercnn} can also be observed by comparing between settings (2), (4), (8). This demonstrates the surprising effectiveness of the proposed Texture Jitter. Settings (5), (6), (7) are designed to verify the effectiveness of our DAF framework. By adding single-center-loss~\cite{SGL} to setting (5), setting (6) gets only tiny improvements. However, by equipping setting (5) with our DAF framework, setting (7) achieves much larger improvements. Setting (7) and setting (8) show that our methods are robust to different base detectors, and that a better base detector can mostly lead to better performance.

In Figure~\ref{fig: matrics}, each matrix has 9$\times9$=81 cells, corresponding to the F-scores of the 81 test settings on the OSFT dataset. Different rows of the matrices denote different training sets and different columns denote different test sets. Darker color indicates better performance. In Figure~\ref{fig: matrics}, the top-left matrix is the result of the baseline model (setting (1) of Table~\ref{tab: OSTFdetcomp}), this matrix shows an obvious diagonal, indicating that the common model only performs well when tested on seen forgeries. For example, in the top-left matrix, the model trained on the text tampered by `STEFANN` gets 85.14 F-score when tested on `STEFANN` (row 2, column 2), but its F-score drops to 0 when tested on the unseen forgery tampered by `SRNet` (row 2, column 1). This poor open-set performance issue is not addressed when the model is pre-trained on common text detection task with the same datasets and configuration as ours but without our Texture Jitter (setting (3)). In contrast, the two matrices at the bottom of Figure~\ref{fig: matrics} are the results of our methods, these matrices have much less obvious diagonal, much more uniform and darker colors, indicating that the models trained with our methods gain much better open-set generalization ability. The qualitative results for visual comparison are shown in appendix, which further validate the proposed methods.

\smallskip
\noindent \textbf{Discussion}. In Figure~\ref{fig: matrics}, models have relatively worse open-set performance when being tested on `STEFANN` and `TextOCR`. for the following reasons: (1) For STEFANN, there is a huge difference in texture appearance between the output of STEFANN and other editing methods. Texts edited by STEFANN have an almost binary texture appearance, while texts edited by other methods have richer and more realistic texture details (Figure~\ref{fig: OTSE}). Therefore, generalization is much more challenging. (2) For TextOCR, it has many tiny and fuzzy texts that are challenging to detect. When models are trained on other source data, they are adapted to big clear text and thus have performance degradation on such tiny and fuzzy texts.
The tiny and fuzzy texts are also the main reason for false positives in other subsets. Despite these challenges, our proposed methods still significantly improve the model.

\section{Conclusion}
In the era of generative AI, this paper introduces a novel task of open-set tampered scene text detection, designed to meet the demands for generalized forensics analysis. This task challenges forensics models with both seen and previously unseen forgeries, aiming to thoroughly evaluate their generalization. To facilitate this, we have manually developed a comprehensive high-quality benchmark, named OSTF, which stands out by including text images tampered by various editing models. Further, we introduce an innovative pre-training paradigm, termed Texture Jitter, which effectively mitigates the lack of high-quality training data and significantly enhances the model capabilities in fine-grained perception and open-set generalization. The proposed Texture Jitter is the first fine-grained perception pre-training paradigm, which differs from previous works that are coarse-grained perception. To further enhance open-set generalization, we also present a novel pluggable framework DAF, which focuses on identifying the feature difference between authentic and tampered texts. This differs from previous works that rely only on the specific input features. Extensive experiments validate the effectiveness of our methods. With these advances, our work achieves a significant step forward in the field of tampered text detection.

\section{Acknowledgment}
This research is supported in part by the National Natural Science Foundation of China (Grant No.: 62441604, 62476093) and IntSig-SCUT Joint Lab Foundation.

\bibliography{aaai25}

\appendix
\input{supplement}

\end{document}

%% file: supplement.tex
\newpage

\section{Supplement}

\begin{abstract}
In this supplementary material, we first provide more details about the proposed Texture Jitter and the DAF framework, we then provide further discussions, more experimental results and qualitative visualization.
\end{abstract}

%

\section{More Details about the Texture Jitter}
The texture processing operations we consider include random blur, reverse blur, random image compression and reverse image compression. More specifically, the random blur operation is randomly selected from (Gaussian Blur, Down-sampling Blur and Motion Blur), the random compression function is implemented by JPEG compression, the reverse blur is achieved by filtering the input image with a sharpening kernel [[0, 1, 0], [1, 5, 1], [0, 1, 0]], and the reverse compression is implemented by a specific algorithm~\cite{FBCNN}.

\begin{figure}[t!]
  \centering
  \includegraphics[width=\linewidth]{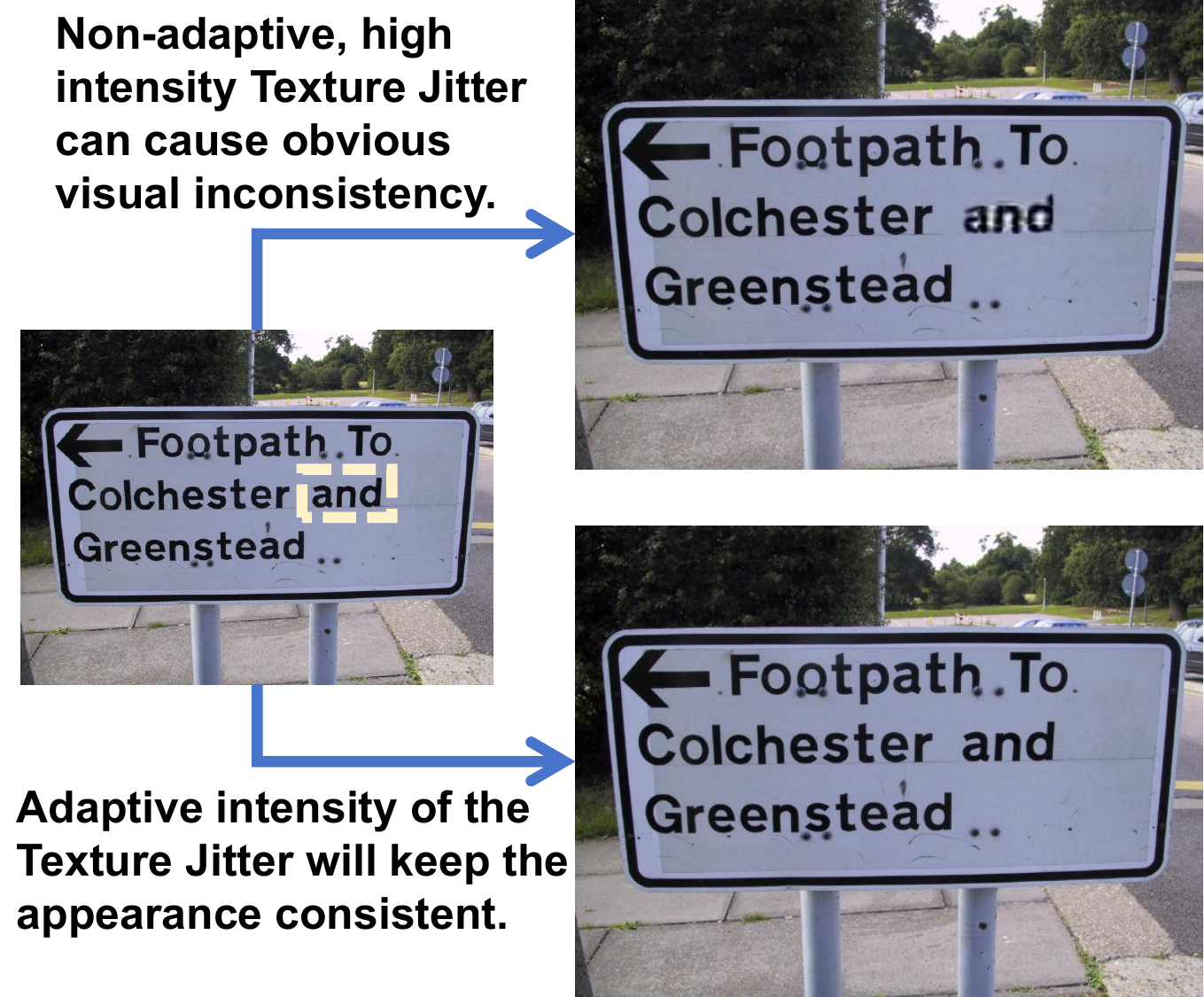}
  \caption{Non-adaptive Texture Jitter can often cause too obvious visual inconsistency in the tampered text (e.g., the text 'and' in the top-right image).}
  \label{fig: nonad_TJ}
\end{figure}

Additionally, we empirically find that the texts of different sizes correspond to different optimal intensity ranges of the Texture Jitter. For example, if a low intensity of Texture Jitter is applied to large text, the processed text will be too similar to the original text. The anomaly becomes too obscure and the model is confused and difficult to converge. On the contrary, if a high intensity of Texture Jitter is applied to small text, the visual inconsistency becomes too strong, as shown in Figure~\ref{fig: nonad_TJ}. Consequently, the model can easily find a shortcut to learn such strong patterns and thus has a notably worse generalization (overfitting to detect obvious anomalies such as the text 'and' in Figure~\ref{fig: nonad_TJ}, and being unable to detect the unobvious anomalies in real-life). Furthermore, to ensure a natural fusion between the original image and the processed regions, the optimal strength of the edge smoothing also differs for different texts. We adaptively adjust the intensity of the Texture Jitter and the edge smoothing with rules based on the size of the target text. The implementation will be open-source. With the proposed adaptive intensity, our Texture Jitter can always produce high-quality outputs in various scenarios.

\section{More Details about the DAF Framework}
\noindent \textbf{Feature extraction.} Given an input image $I$, a backbone model $B$ is used to extract the encoder's output feature $E$ as $E=B(I)$. The multi-scale features $F_{box}$ for common text detection and $F_{cls}$ for text forensics are extracted by two Feature Pyramid Networks $FPN_1$ and $FPN_2$ separately as $F_{box}=FPN_1(E)$, $F_{cls}=FPN_2(E)$. 

\smallskip

\noindent \textbf{Text Detection.} Given the input multi-scale features $F_{box}$, the text box localization predictions $P_{box}$ are obtained by $P_{box}=Det(F_{box})$, where $Det$ denotes the same detection modules (RPN and RoI Head) in Faster R-CNN. 

\smallskip

\noindent \textbf{Text Forensics.} Given the input multi-scale features $F_{cls}$ and text localization predictions $P_{box}$, the feature vector $V_{cls}$ of each text in the forensics branch is obtained by $V_{cls}=RoIAlign(F_{cls}, P_{box})$. The modulated authentic kernel vector $V_m=FC((Avg(F_{cls}), V_{cls}))$, where $Avg$ is global average pooling and $FC$ denotes fully-connected layers. The final prediction $P_{cls}$ for forensics (real / fake classification) is obtained by $P_{cls}=FC(V_{cls}-V_m)$

\smallskip

\noindent \textbf{Loss Functions.} The proposed DAF framework is trained in an end-to-end manner with the loss function $L_{all}$:

\begin{equation}
    L_{all} = L_{cls} + L_{bbox} + L_{feat}
    \label{1}
\end{equation}

In equation~(1), $L_{cls}$ is the binary cross-entropy loss for authentic/tampered text classification:

\begin{equation}
    L_{cls} = y \times log(pred) + (1 - y) \times log(1 - pred)
    \label{2}
\end{equation}

In equation~(2), `pred` is the model prediction and `y` is the ground-truth classification label.

In equation~(1), $L_{bbox}$ is the L1 loss for bounding box regression, which is the same as that in Faster-RCNN~\cite{ren2015fasterrcnn} and Cascade-RCNN~\cite{cai2018cascadercnn}.

In equation~(1), $L_{feat}$ is the loss function for learning the authentic kernel vector:

\begin{equation}
    L_{feat} = Dist_{auth} + Max(Dist_{auth} - Dist_{tamp} + 32, 0)
    \label{3}
\end{equation}

In equation~(3), $Dist_{auth}$ is the distance between the authentic kernel and the RoI vectors of the authentic texts:

\begin{equation}
    Dist_{auth} = \frac{1}{N_{auth}}\sum\limits_{i\in N_{auth}}^{} = ||a_{i} - K||_{2}
    \label{4}
\end{equation}

In equation~(4), $a_{i}$ is the $i_{th}$ RoI vector of the authentic text, $K$ is the authentic kernel.

In equation~(3), $Dist_{tamp}$ is the distance between the authentic kernel and the RoI vectors of the tampered texts:

\begin{equation}
    Dist_{tamp} = \frac{1}{N_{tamp}}\sum\limits_{i\in N_{tamp}}^{} = ||t_{i} - K||_{2}
    \label{5}
\end{equation}

In equation~(5), $t_{i}$ is the $i_{th}$ RoI vector of the tampered text, $K$ is the authentic kernel.

During inference, the proposed DAF framework still works in an E2E manner with the learned authentic kernel, as shown in Figure~5 of the paper.

\section{More Implement Details}
\textbf{Converting Box Predictions to Segmentation Maps}. For a fair comparison with the segmentation-based methods, we convert the bounding box predictions of our model into segmentation maps. To achieve this, we first initialize two zero maps of the same resolution with the input image, one for authentic text and one for tampered text. We then highlight the predicted tampered text box regions on the zero map for tampered text and similar to the predicted real text. With the obtained segmentation maps, we calculate the IoU and F1-score at the pixel-level with the built-in functions of mmsegmentation~\cite{contributors2020mmsegmentation}

\noindent \textbf{Base detectors}. For the base detector, we adopt the Faster R-CNN~\cite{ren2015fasterrcnn} and Cascade R-CNN~\cite{cai2018cascadercnn} that implemented in the mmdetection~\cite{chen2019mmdetection} library. In each feature map of stride [4, 8, 16, 32, 64], the base anchor size is set to 8$\times$stride and the height/width ratios are set to [0.25, 0.5, 1.0]. 

\section{Discussions}
\begin{figure}[t!]
  \centering
  \includegraphics[width=0.95\linewidth]{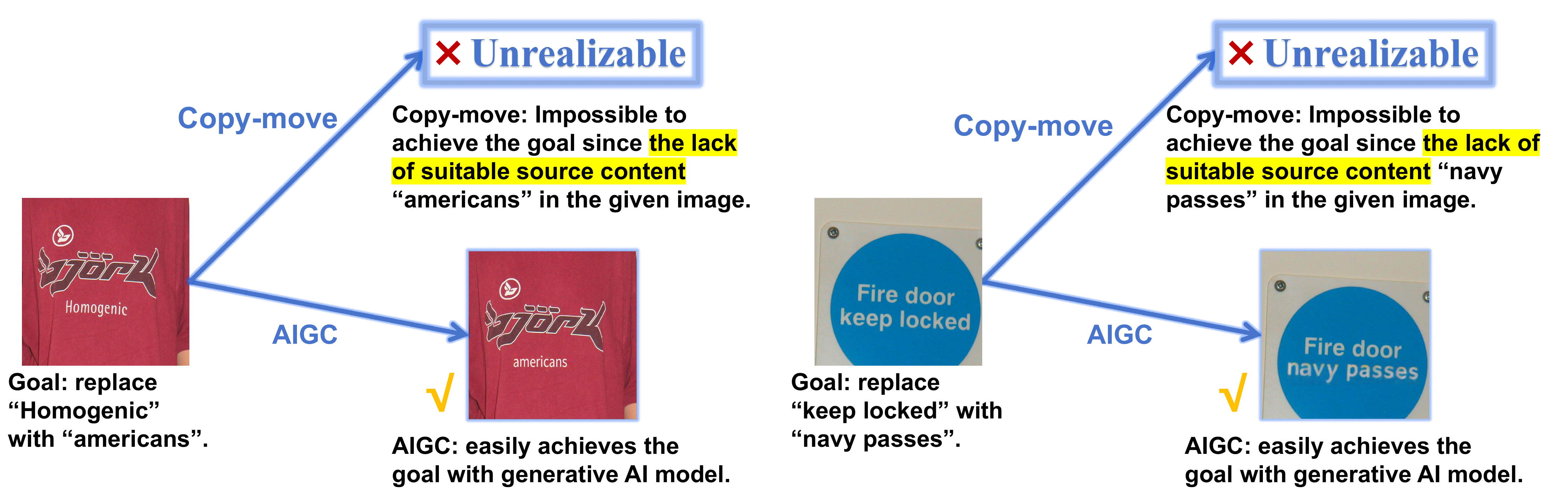}
  \caption{Copy-move is rarely applicable in real-world scenarios due to the lack of a suitable source region.}
  \label{fig: copymove}
\vspace{-0.2cm}
\end{figure}

\noindent \textbf{Copy-move is rarely applicable to scene text editing}. 
Copy-move requires suitable source region and is rarely applicable to our focus, scene text, due to the complex style and sparseness of scene text~\cite{wang2022tic13}. For example, in the two cases of Figure~\ref{fig: copymove}, it's impossible to find any region containing target text (e.g., `americans`) in the given image. In contrast, AI-based text editing has fewer constrains, is much easier to use, and is therefore much more common for scene text in real life. We therefore focus on detecting generative forgery for scene text.

\noindent \textbf{The selection of the source data}.
It's not feasible to use some text image data sources (ReCTS, ICDAR2015, ICDAR2017, ArT) for dataset construction, since existing text editing models only work well for English, straight, and clear text. Unfortunately, most of ReCTS~\cite{rects} and ICDAR2017~\cite{ic17mlt} are non-English text, ArT~\cite{art} is curve text, and ICDAR2015~\cite{ic15} is too blurry text. Current text editing models can hardly modify them into visually satisfactory content. Despite the challenge, to enrich the source data for text tampering, we still manually pick the best images from ReCTS and ICDAR2017, and manually forge them using Textdiffuser~\cite{textdiffuserv2}.

\noindent \textbf{Why the zero-shot version of our method outperforms previous full-shot SOTA}. Due to the costly production of manually refined data, the high-quality training data in real-life is of tiny size (e.g. only 200 tampered images in the training set of Tampered-IC13). This leads to overfitting of most the deep models~\cite{wang2022tic13}. However, by applying our Texture Jitter to various text images and pre-training the model to detect the processed text, the pre-trained model can accurately identify the obscure texture anomaly left by the editing model. Therefore, the zero-shot version of our method achieves better generalization.

\noindent \textbf{Effects of the base detector}.
Our work focuses on detecting the text that was tampered by sophisticated methods, with text detection as a base module. Based on extensive experiments, our methods are robust to different base text detectors, while a better base detector will produce stronger results. To further analyze the effects of the base detector, we replace the outputs of the base text detection branch with ground-truth bounding boxes and evaluate model performance in an ideal scenario. The results on the Tampered-IC13 benchmark~\cite{wang2022tic13} and the OSFT benchmark are shown in Table~\ref{tab: TIC Perfect} and Table~\ref{tab: OSFT Perfect} respectively. With the ideal detector, the model performance has a significant gain. The performance gap between the different models becomes smaller, suggesting that the text detection branch and the text forensics branch work independently.

\noindent \textbf{It is not practical to train directly on purely generated data}.
To further validate the effectiveness of our proposed Texture Jitter, and the statement in the pioneering work~\cite{wang2022tic13} that the poor performance is caused by the training data generated by the existing text editing models without any manual intervention, we pre-train the models on the purely generated data. We select the current text editing methods from each of the three types, respectively (conventional, rendering-based and diffusion-based), namely MOSTEL~\cite{AAAI_MOSTEL}, DST~\cite{Shimoda_2021_ICCV} and AnyText~\cite{Anytext}. We edit the text in the same pre-training datasets as our Texture Jitter, including LSVT~\cite{lsvt}, ReCTS~\cite{rects}, ICDAR2013~\cite{karatzas2013icdar}, ICDAR2015~\cite{ic15}, ICDAR2017~\cite{ic17mlt}, TextOCR~\cite{singh2021textocr}, ArT~\cite{art}, and train our model with the edited text images and the same training configuration. Based on experiments, we find that these data will always lead to a NAN result at the early training stage and the model will always fail to converge. By performing visualization and detailed analysis, we figure out the reason that the generated data are too noisy. When MOSTEL~\cite{AAAI_MOSTEL} and DST~\cite{Shimoda_2021_ICCV} encounter the text with an unfamiliar, complex style, they cannot perceive the text and the processed texts are not edited and almost the same as the original ones, while the processed texts have labels of 'fake'. These cases are difficult to filter out automatically, since the output will be different from the original due to the resizing operation. The AnyText~\cite{Anytext} sometimes just erases the target region and does not produce any new text on it, but the target region will still have a label of `fake box`. As a result, the produced data is too noisy for the model to learn meaningful knowledge. \textbf{These experiments prove that manual improvement in post-processing is necessary, which is consistent with the results of the pioneering work and further validates the advantages of our Texture Jitter}.

\noindent \textbf{Differences between DAF and anchor-based losses}. Although both of DAF and anchor-based loss cluster features during training, our DAF clearly differs from anchor-based loss by explicitly modeling the feature difference, feeding the classifier with the subtraction between the input features and the authentic kernel, rather than just feeding the classifier with the input features or manual threshold filtering as what anchor-based losses do (e.g. triplet loss).

\section{Extensive Ablation Study on the Tampered-IC13 Dataset} The ablation results on the Tampered-IC13 dataset are shown in Table~\ref{tab: ticdetabl}. where both of the proposed Texture Jitter and DAF framework contribute towards higher performance.

\section{Extensive Comparison Study \\ with AIGC-Detection Models}
In this section, we conduct an extra comparison with the methods from a related field, AI-generated image detection, to further validate the effectiveness of the proposed methods.

\noindent \textbf{Implementation Details}. We select two SOTA AI-generated image detection methods, including UniFD~\cite{UniFD} and SSP~\cite{niuli2024single}. The selected methods have achieved significant progress on the widely-used AI-generated image detection benchmarks. We implement the UniFD~\cite{UniFD} with its official code. Considering that these methods only identify an image as fake when it is \textbf{generated as a whole}, we crop the text regions based on their ground-truth bounding boxes, resize the cropped texts into the required size, and feed them into the binary classification model. These two models are trained on the OSFT dataset using their official training configurations and the same batch size and iterations as ours.

\noindent \textbf{Evaluation Metrics}. We adopt instance-level Precision (P), Recall (R) and F1-score (F) as evaluation metrics.
The inputs are texts cropped by the ground-truth boxes. For a fair comparison, we also use our `perfect detector` models.

\noindent \textbf{Results}. The results on the Tampered-IC13 benchmark are shown in Table~\ref{tab: comp OSFT AIGCD}. Our model significantly outperforms both SOTA models on tampered text detection. It's notable that the outperformance of our model is achieved with a much smaller input size and much fewer model parameters, as shown in the right of the Table~\ref{tab: comp OSFT AIGCD}.

\begin{figure}[t!]
  \centering
  \includegraphics[height=110pt, width=0.9\linewidth]{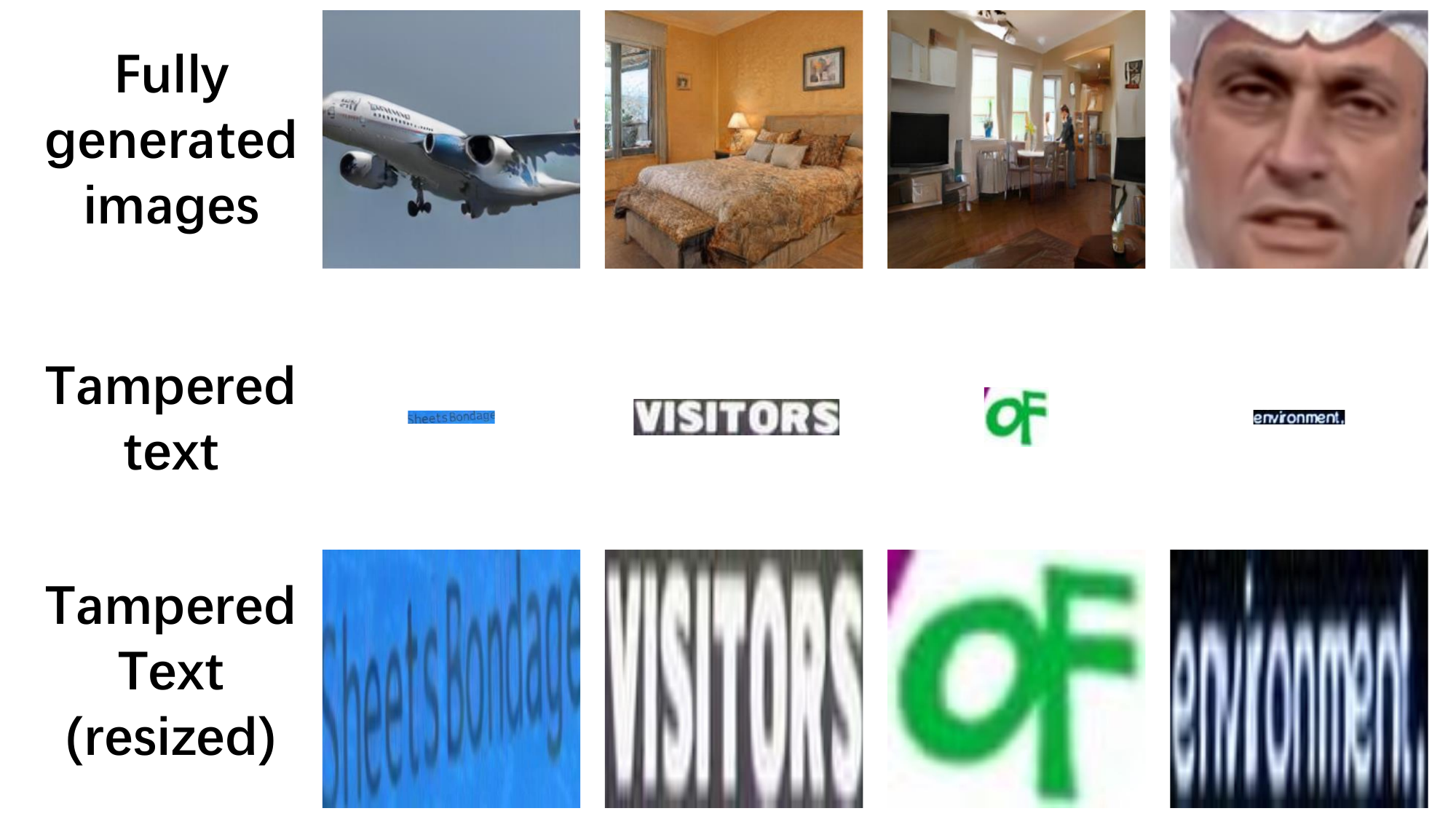}
  \caption{The fully generated images in the widely-used AI generated image detection benchmarks have significant domain gaps between tampered text.}
  \label{fig: supple}
\end{figure}

\noindent \textbf{Analysis}. Although the two selected models are good at detecting fully generated images, the cropped tampered texts have significant differences between the image-level forgeries in many aspects, such as style and resolution, as shown in Figure~\ref{fig: supple}. These gaps prevent the core designs of the AI-generated image recognition models from working well (e.g. the fixed CLIP~\cite{clip} pre-trained weights in UniFD~\cite{UniFD} do not fit with the image style of the tampered text).

\section{Robustness Evaluation}

Following the previous work~\cite{wang2022tic13}, we set the input image size to `short side less than 1200, long side less than 2000`, and evaluate model performance under compression and zoom distortion. The results on the Tampered-IC13~\cite{wang2022tic13} benchmark are shown in Table~\ref{tab: TIC Robust}. Our model demonstrates satisfactory robustness under different distortions.

\begin{figure*}[!ht]
  \centering
  \includegraphics[width=\linewidth]{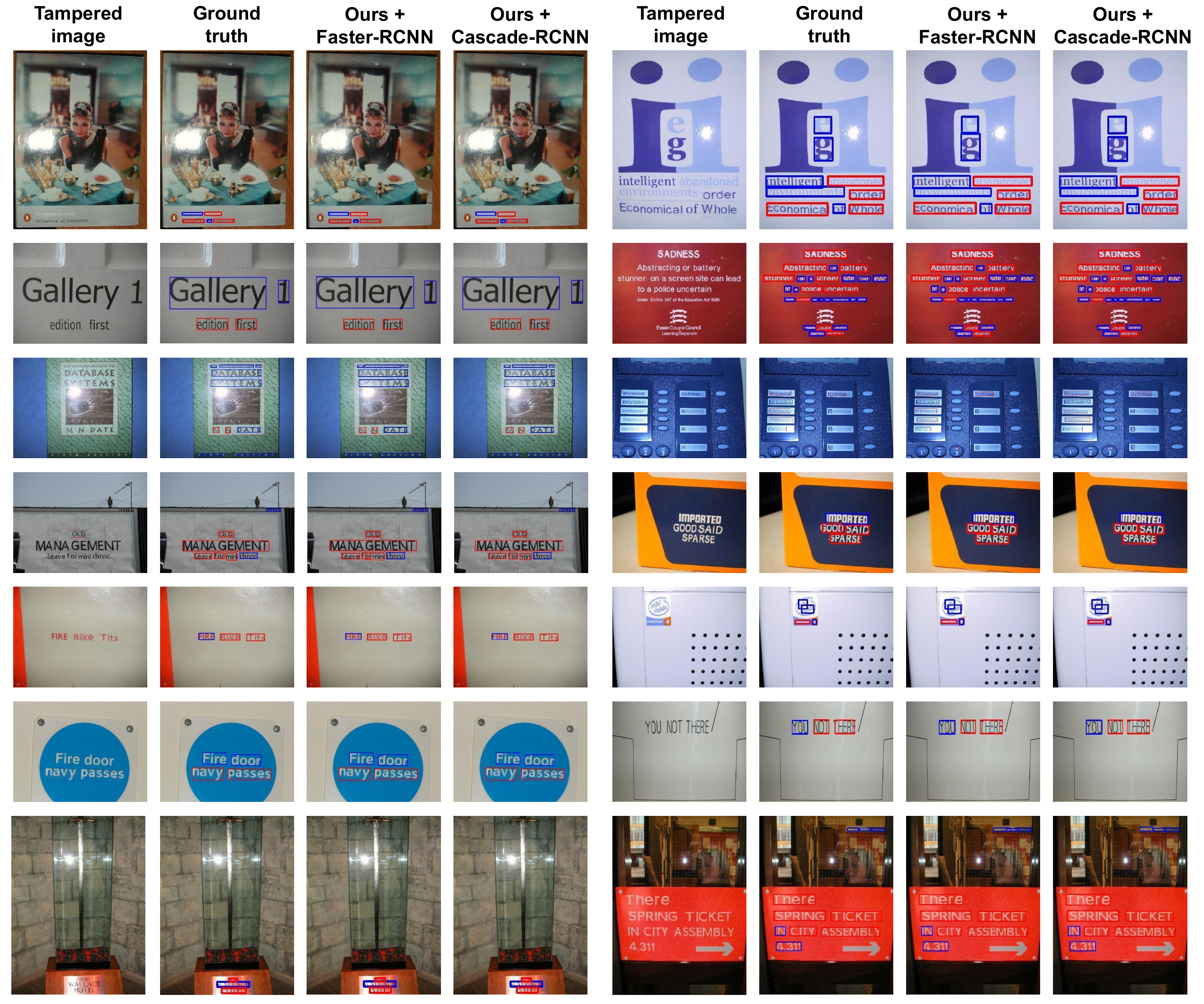}
  \caption{Visualization. Blue boxes for real text and red boxes for tampered text.}
  \label{fig: sp2}
\end{figure*}

\section{Comparison with DocTamper dataset}
The comparison between our OSFT dataset and the DocTamper dataset~\cite{CVPR2023DocTamper} is shown in Table~\ref{tab: doctamper}. Our dataset stands out by including 100\% handmade forgeries and text tampered by various AI models, while the DocTamper dataset has neither.

\section{Ablation study for noise features}
The ablation study for noise features on the OSTF dataset is shown in Table~\ref{tab: abl_noise}. The models with different extra input noise features as input perform slightly worse than the original pure RGB model (RGB+None) on the OSTF dataset. 

\section{Comparison study for anchor-based loss}
The comparison study for anchor-based loss on the OSTF dataset is shown in Table~\ref{tab: comp_loss}. The proposed DAF achieves better performance than Triplet loss, SCL, Contrast loss.

\begin{table}[t!]
\vspace{-0.2cm}
\caption{Comparison with the DocTamper dataset.}
\vspace{-0.3cm}
\setlength{\tabcolsep}{1pt}
\begin{tabular}{ccccccccc}
\hline
Dataset &  & {\makecell{Handmade\\data}} &  & {\makecell{Synthetic\\data}} &  & {\makecell{Forged by\\AI models}} &  & {\makecell{Types of\\AI models}} \\ \cline{1-1} \cline{3-3} \cline{5-5} \cline{7-7} \cline{9-9} 
DocTamper &  & 0 &  & \textbf{17k} &  & 0 &  & 0 \\ \cline{1-1} \cline{3-3} \cline{5-5} \cline{7-7} \cline{9-9} 
Ours &  & \textbf{1980} &  & 0 &  & \textbf{100\%} & \textbf{} & \textbf{8} \\ \hline
\end{tabular}
\label{tab: doctamper}
\end{table}

\begin{table}[!t]
\caption{Ablation for the noise features on OSTF.}
\vspace{-0.3cm}
\centering
\begin{tabular}{ccccc}
\hline
RGB+ & None          & SRM  & Bayar & NoisePrint++ \\ \hline
Mean F1   & \textbf{74.8} & 74.8 & 73.8  & 74.4         \\ \hline
\end{tabular}
\label{tab: abl_noise}
\end{table}

\begin{table}[!t]
\small
\caption{Comparison for loss functions on OSTF.}
\vspace{-0.3cm}
\centering
\begin{tabular}{ccccc}
\hline
Loss & DAF (Ours)          & Contrast loss  & Triplet loss & SCL \\ \hline
Mean F1   & \textbf{74.8} & 73.9 &  73.6  & 73.8         \\ \hline
\end{tabular}
\label{tab: comp_loss}
\end{table}

\begin{table*}[h]
\caption{Ablation study on the Tampered-IC13 dataset. `P`, `R`, `F`, `mF` denote precision, recall, F1-score and mean F1-score respectively. `SCL` denotes the use of single-center-loss~\cite{SGL}. `DAF` denotes our Difference-Aware Forensics.}
\vspace{-0.35cm}
\setlength{\tabcolsep}{1.5pt}
\begin{tabular}{ccccccccccccccccccccccccccccc}
\hline
\multicolumn{15}{c}{Ablation settings} &  & \multicolumn{13}{c}{Evaluation Score} \\ \cline{1-15} \cline{17-21} \cline{23-27} \cline{29-29} 
\multirow{2}{*}{\makecell{Num\\ber}} &  & \multicolumn{5}{c}{Pre-training} &  & \multicolumn{3}{c}{Framework} &  & \multicolumn{3}{c}{Base detector} &  & \multicolumn{5}{c}{Real Text} &  & \multicolumn{5}{c}{Tampered Text} &  & Average \\ \cline{3-7} \cline{9-11} \cline{13-15} \cline{17-21} \cline{23-27} \cline{29-29} 
 &  & \makecell{COCO\\Detection} &  & \makecell{Text\\Detection} &  & \makecell{Texture\\Jitter (Ours)} &  & \makecell{SCL} &  & \makecell{DAF\\(Ours)} &  & \makecell{Faster\\RCNN} &  & \makecell{Cascade\\RCNN} &  & P &  & R &  & F &  & P &  & R &  & F &  & mF \\ \cline{1-1} \cline{3-3} \cline{5-5} \cline{7-7} \cline{9-9} \cline{11-11} \cline{13-13} \cline{15-15} \cline{17-17} \cline{19-19} \cline{21-21} \cline{23-23} \cline{25-25} \cline{27-27} \cline{29-29} 
(1) &  & \checkmark &  &  &  &  &  &  &  &  &  & \checkmark & \textbf{} &  &  & 68.77 &  & 66.39 &  & 67.56 & \textbf{} & 85.34 &  & 94.26 &  & 89.58 & \textbf{} & 78.57 \\
(2) &  & \checkmark &  &  &  &  &  &  &  & \checkmark &  &  & \textbf{} & \checkmark &  & 70.92 &  & 68.01 &  & 69.43 & \textbf{} & 86.06 &  & 96.25 &  & 90.87 & \textbf{} & 80.15 \\
(3) &  &  &  & \checkmark &  &  &  &  &  &  &  & \checkmark &  &  &  & 87.25 &  & 72.16 &  & 78.99 &  & 86.50 &  & 97.13 &  & 91.51 &  & 85.25 \\
(4) &  &  &  & \checkmark &  &  &  &  &  & \checkmark &  &  &  & \checkmark &  & \textbf{89.17} &  & 73.42 &  & 80.53 &  & 88.68 &  & \textbf{97.22} &  & 92.75 &  & 86.64 \\
(5) &  &  &  &  &  & \checkmark &  &  &  &  &  & \checkmark &  &  &  & 78.37 &  & 82.37 &  & 80.32 &  & 89.50 &  & 96.11 &  & 92.69 &  & 86.51 \\
(6) &  &  &  &  &  & \checkmark &  & \checkmark &  &  &  & \checkmark &  &  &  & 78.71 &  & 82.21 &  & 80.42 &  & 90.17 &  & 95.90 &  & 92.95 &  & 86.67 \\
(7) &  &  &  &  &  & \checkmark &  &  &  & \checkmark &  & \checkmark &  &  &  & 80.52 &  & 81.71 &  & 81.11 &  & 91.44 &  & 96.31 &  & 93.81 &  & 87.46 \\
(8) &  &  &  &  &  & \checkmark &  &  &  & \checkmark &  &  &  & \checkmark &  & 82.95 &  & \textbf{82.54} &  & \textbf{82.74} &  & \textbf{92.37} &  & 96.72 &  & \textbf{94.49} &  & \textbf{88.62} \\ \hline
\end{tabular}
\label{tab: ticdetabl}
\end{table*}

\begin{table*}[h!]
\caption{Ablation study of the base detector on the Tampered-IC13 dataset, the model performance is evaluated with the official detection-based metrics at the instance-level. `P`, `R`, `F`, `mF` denote precision, recall, F1-score and mean F1-score.}
\setlength{\tabcolsep}{5pt}
\begin{tabular}{ccccccccccccccc}
\hline
{}                                    & {} & \multicolumn{5}{c}{{Real Text}}                                                                                           & {} & \multicolumn{5}{c}{{Tampered Text}}                                                                                       & {} & {Average} \\ \cline{3-7} \cline{9-13} \cline{15-15} 
\multirow{-2}{*}{{Method}}            & {} & {P}     & {} & {R}     & {} & {F}     & {} & {P}     & {} & {R}     & {} & {F}     & {} & {mF}      \\ \cline{1-1} \cline{3-3} \cline{5-5} \cline{7-7} \cline{9-9} \cline{11-11} \cline{13-13} \cline{15-15} 
{ours+Faster-RCNN}                     & {} & {80.52} & {} & {81.71} & {} & {81.11} & {} & {91.44} & {} & {96.31} & {} & {93.81} & {} & {87.46}   \\
{ours+Cascade-RCNN}                    & {} & {82.95} & {} & {82.54} & {} & {82.74} & {} & {92.37} & {} & {96.72} & {} & {94.49} & {} & {88.62}   \\
{ours+Faster-RCNN (Perfect Detector)}  & {} & \textbf{99.32} & {} & \textbf{96.71} & {} & \textbf{98.00} & {} & \textbf{96.99} & {} & {98.98} & {} & \textbf{97.97} & {} & \textbf{97.98}   \\
{ours+Cascade-RCNN (Perfect Detector)} & {} & \textbf{99.32} & {} & {95.72} & {} & {97.48} & {} & {96.41} & {} & \textbf{99.18} & {} & {97.78} & {} & {97.63}   \\ \hline
\end{tabular}
\label{tab: TIC Perfect}
\end{table*}

\begin{table*}[h!]
\caption{Ablation study of the base detector on the OSFT dataset, the model performance is evaluated using detection-based metrics at the instance-level. `mP`, `mR`, `mF` denote mean precision, mean recall and mean F1-score respectively.}
\setlength{\tabcolsep}{5pt}
\begin{tabular}{ccccccccccccccc}
\hline
{ }                                    & { } & \multicolumn{5}{c}{{ Real Text}}                                                                                           & { } & \multicolumn{5}{c}{{Tampered Text}}                                                                                       & { } & { Average} \\ \cline{3-7} \cline{9-13} \cline{15-15} 
\multirow{-2}{*}{{ Method}}            & { } & {mP}     & { } & {mR}     & { } & {mF}     & { } & {mP}     & { } & {mR}     & { } & {mF}     & { } & { mF}      \\ \cline{1-1} \cline{3-3} \cline{5-5} \cline{7-7} \cline{9-9} \cline{11-11} \cline{13-13} \cline{15-15} 
{ ours+Faster-RCNN}                     & { } & 78.54 &  & {76.55} &  & 75.98 &  & 77.55 &  & 72.62 &  & 73.66 &  & 74.82   \\
{ ours+Cascade-RCNN}                    & { } & {82.54} &  & 74.63 &  & {76.74} &  & {80.42} &  & {72.43} &  & {74.96} &  & {75.85}   \\
{ ours+Faster-RCNN (Perfect Detector)}  & { } & \textbf{93.11} & { } & \textbf{95.75} & { } & \textbf{95.43} & { } & {82.49} & { } & {76.34} & { } & {76.74} & { } & \textbf{86.09}   \\
{ ours+Cascade-RCNN (Perfect Detector)} & { } & {92.89} & { } & {95.01} & { } & {95.01} & { } & \textbf{82.85} & { } & \textbf{76.41} & { } & \textbf{76.89} & { } & {85.95}   \\ \hline
\end{tabular}
\label{tab: OSFT Perfect}
\end{table*}

\begin{table*}[h!]
\caption{Comparison study on the Tampered-IC13 dataset, the model performance is evaluated at the instance-level. `mP`, `mR`, `mF` denote mean precision, mean recall and mean F1-score respectively.  The perfect text detector is used for all of the models. `Size` and `Params` denote the input size and the model parameters of the forensics branch respectively.}
\setlength{\tabcolsep}{3pt}
\begin{tabular}{ccccccccccccccccccc}
\hline
{ }                         & { } & \multicolumn{5}{c}{{ Real Text}}                                                                                                                      & { } & \multicolumn{5}{c}{{ Tampered Text}}                                                                                                                  & { } & { Average}        &  &                            &  &                             \\ \cline{3-7} \cline{9-13} \cline{15-15}
\multirow{-2}{*}{{ Method}} & { } & { mP}             & { } & { mR}             & { } & { mF}             & { } & { mP}             & { } & { mR}             & { } & { mF}             & { } & { mF}             &  & \multirow{-2}{*}{Size} &  & \multirow{-2}{*}{Params} \\ \cline{1-1} \cline{3-3} \cline{5-5} \cline{7-7} \cline{9-9} \cline{11-11} \cline{13-13} \cline{15-15} \cline{17-17} \cline{19-19} 
{ SSP~\cite{niuli2024single}}                      & { } & { 72.78}          & { } & { 71.94}          & { } & { 70.94}          & { } & { 34.12}          & { } & { 34.57}          & { } & { 29.23}          & { } & { 50.09}          &  & 256$\times$256                    &  & 23.5M                       \\
UniFD~\cite{UniFD}                                           &                         & 89.14                                 &                         & 78.39                                 &                         & 82.18                                 &                         & 39.10                                 &                         & 54.44                                 &                         & 38.82                                 &                         & 60.50                                 &  & 224$\times$224                    &  & 427.6M                      \\
{ ours+Faster-RCNN}         & { } & { \textbf{92.27}} & { } & { \textbf{95.86}} & { } & { \textbf{95.21}} & { } & { \textbf{83.62}} & { } & { \textbf{73.74}} & { } & { 75.93}          & { } & { \textbf{85.57}} &  & \textbf{8$\times$8}                        &  &                \textbf{6.5M}             \\
{ ours+Cascade-RCNN}        & { } & { 92.02}          & { } & { 95.14}          & { } & { 94.80}          & { } & { 80.70}          & { } & { \textbf{73.74}} & { } & { \textbf{76.17}} & { } & { 85.48}          &  & \textbf{8$\times$8}                        &  &              \textbf{6.5M}              \\ \hline
\end{tabular}
\label{tab: comp OSFT AIGCD}
\end{table*}

\vspace{+0.1cm}
\begin{table*}[h!]
\caption{Robustness evaluation on the Tampered-IC13 dataset, the model performance is evaluated at the pixel-level. `P`, `R`, `F` denote precision, recall and F1-score respectively. `JPEG75` denotes re-compression of the image using jpeg compression with a quality factor of 75. `Resize0.5` denotes the resizing of the image to halve its height and width.}
\begin{tabular}{ccccccccccccccc}
\hline
{ }                             & { } & \multicolumn{5}{c}{{ Real Text}}                                                                                                                                               & { } & \multicolumn{5}{c}{{Tampered Text}}                                                                                                                                           & { } & { Average} \\ \cline{3-7} \cline{9-13} \cline{15-15} 
\multirow{-2}{*}{{ Method}}     & { } & \multicolumn{1}{c}{{ P}} & \multicolumn{1}{c}{{ }} & \multicolumn{1}{c}{{ R}} & { } & { F}     & { } & \multicolumn{1}{c}{{ P}} & \multicolumn{1}{c}{{ }} & \multicolumn{1}{c}{{ R}} & { } & { F}     & { } & { mF}      \\ \cline{1-1} \cline{3-3} \cline{5-5} \cline{7-7} \cline{9-9} \cline{11-11} \cline{13-13} \cline{15-15} 
{ ours+Cascade-RCNN (Original)}  & { } & { 86.14}                 & { }                     & { 92.05}                 & { } & { 88.99} & { } & { 94.88}                 & { }                     & { 94.34}                 & { } & { 94.61} & { } & { 91.80}   \\
{ ours+Cascade-RCNN (JPEG75)}    & { } & { 88.70}                 & { }                     & { 90.36}                 & { } & { 89.52} & { } & { 93.14}                 & { }                     & { 92.19}                 & { } & { 92.66} & { } & { 91.09}   \\
{ ours+Cascade-RCNN (Resize0.5)} & { } & { 88.64}                 & { }                     & { 94.05}                 & { } & { 91.26} & { } & { 93.83}                 & { }                     & { 93.13}                 & { } & { 93.48} & { } & { 92.37}   \\ \hline
\end{tabular}
\label{tab: TIC Robust}
\end{table*}

\bigskip

~\\

%% file: anonymous-submission-latex-2025.bbl
\begin{thebibliography}{50}
\providecommand{\natexlab}[1]{#1}

\bibitem[{Bao et~al.(2021)Bao, Dong, Piao, and Wei}]{bao2021beit}
Bao, H.; Dong, L.; Piao, S.; and Wei, F. 2021.
\newblock Beit: Bert pre-training of image transformers.
\newblock \emph{arXiv preprint arXiv:2106.08254}.

\bibitem[{Cai and Vasconcelos(2018)}]{cai2018cascadercnn}
Cai, Z.; and Vasconcelos, N. 2018.
\newblock Cascade r-cnn: Delving into high quality object detection.
\newblock In \emph{Proceedings of the IEEE conference on computer vision and pattern recognition}, 6154--6162.

\bibitem[{Chen et~al.(2023{\natexlab{a}})Chen, Huang, Lv, Cui, Chen, and Wei}]{textdiffuserv2}
Chen, J.; Huang, Y.; Lv, T.; Cui, L.; Chen, Q.; and Wei, F. 2023{\natexlab{a}}.
\newblock TextDiffuser-2: Unleashing the Power of Language Models for Text Rendering.
\newblock \emph{arXiv preprint arXiv:2311.16465}.

\bibitem[{Chen et~al.(2023{\natexlab{b}})Chen, Huang, Lv, Cui, Chen, and Wei}]{textdiffuser}
Chen, J.; Huang, Y.; Lv, T.; Cui, L.; Chen, Q.; and Wei, F. 2023{\natexlab{b}}.
\newblock TextDiffuser: Diffusion Models as Text Painters.
\newblock In Oh, A.; Neumann, T.; Globerson, A.; Saenko, K.; Hardt, M.; and Levine, S., eds., \emph{Advances in Neural Information Processing Systems}, volume~36, 9353--9387. Curran Associates, Inc.

\bibitem[{Chen, Yao, and Niu(2024)}]{niuli2024single}
Chen, J.; Yao, J.; and Niu, L. 2024.
\newblock A Single Simple Patch is All You Need for AI-generated Image Detection.
\newblock \emph{arXiv preprint arXiv:2402.01123}.

\bibitem[{Chen et~al.(2019)Chen, Wang, Pang, Cao, Xiong, Li, Sun, Feng, Liu, Xu et~al.}]{chen2019mmdetection}
Chen, K.; Wang, J.; Pang, J.; Cao, Y.; Xiong, Y.; Li, X.; Sun, S.; Feng, W.; Liu, Z.; Xu, J.; et~al. 2019.
\newblock MMDetection: Open mmlab detection toolbox and benchmark.
\newblock \emph{arXiv preprint arXiv:1906.07155}.

\bibitem[{Chen et~al.(2018)Chen, Zhu, Papandreou, Schroff, and Adam}]{chen2018deeplabv3+}
Chen, L.-C.; Zhu, Y.; Papandreou, G.; Schroff, F.; and Adam, H. 2018.
\newblock Encoder-decoder with atrous separable convolution for semantic image segmentation.
\newblock In \emph{Proceedings of the European conference on computer vision (ECCV)}, 801--818.

\bibitem[{Chng et~al.(2019)Chng, Liu, Sun, Ng, Luo, Ni, Fang, Zhang, Han, Ding et~al.}]{art}
Chng, C.~K.; Liu, Y.; Sun, Y.; Ng, C.~C.; Luo, C.; Ni, Z.; Fang, C.; Zhang, S.; Han, J.; Ding, E.; et~al. 2019.
\newblock Icdar2019 robust reading challenge on arbitrary-shaped text-rrc-art.
\newblock In \emph{2019 International Conference on Document Analysis and Recognition (ICDAR)}, 1571--1576. IEEE.

\bibitem[{Contributors(2020)}]{contributors2020mmsegmentation}
Contributors, M. 2020.
\newblock MMSegmentation: Openmmlab semantic segmentation toolbox and benchmark.

\bibitem[{Dong et~al.(2023)Dong, Wang, Ji, Liang, Fan, and Ge}]{DFD1}
Dong, S.; Wang, J.; Ji, R.; Liang, J.; Fan, H.; and Ge, Z. 2023.
\newblock Implicit identity leakage: The stumbling block to improving deepfake detection generalization.
\newblock In \emph{Proceedings of the IEEE/CVF Conference on Computer Vision and Pattern Recognition}, 3994--4004.

\bibitem[{Ji et~al.(2023)Ji, Zhang, Wang, Hou, Zhang, Price, and Chang}]{DiffSTE}
Ji, J.; Zhang, G.; Wang, Z.; Hou, B.; Zhang, Z.; Price, B.; and Chang, S. 2023.
\newblock Improving diffusion models for scene text editing with dual encoders.
\newblock \emph{arXiv preprint arXiv:2304.05568}.

\bibitem[{Jiang, Zhang, and Timofte(2021)}]{FBCNN}
Jiang, J.; Zhang, K.; and Timofte, R. 2021.
\newblock Towards flexible blind JPEG artifacts removal.
\newblock In \emph{Proceedings of the IEEE/CVF International Conference on Computer Vision}, 4997--5006.

\bibitem[{Karatzas et~al.(2015)Karatzas, Gomez-Bigorda, Nicolaou, Ghosh, Bagdanov, Iwamura, Matas, Neumann, Chandrasekhar, Lu et~al.}]{ic15}
Karatzas, D.; Gomez-Bigorda, L.; Nicolaou, A.; Ghosh, S.; Bagdanov, A.; Iwamura, M.; Matas, J.; Neumann, L.; Chandrasekhar, V.~R.; Lu, S.; et~al. 2015.
\newblock ICDAR 2015 competition on robust reading.
\newblock In \emph{2015 13th international conference on document analysis and recognition (ICDAR)}, 1156--1160. IEEE.

\bibitem[{Karatzas et~al.(2013)Karatzas, Shafait, Uchida, Iwamura, i~Bigorda, Mestre, Mas, Mota, Almazan, and De~Las~Heras}]{karatzas2013icdar}
Karatzas, D.; Shafait, F.; Uchida, S.; Iwamura, M.; i~Bigorda, L.~G.; Mestre, S.~R.; Mas, J.; Mota, D.~F.; Almazan, J.~A.; and De~Las~Heras, L.~P. 2013.
\newblock ICDAR 2013 robust reading competition.
\newblock In \emph{2013 12th international conference on document analysis and recognition}, 1484--1493. IEEE.

\bibitem[{Li et~al.(2021{\natexlab{a}})Li, Xie, Li, Wang, and Zhang}]{SGL}
Li, J.; Xie, H.; Li, J.; Wang, Z.; and Zhang, Y. 2021{\natexlab{a}}.
\newblock Frequency-aware discriminative feature learning supervised by single-center loss for face forgery detection.
\newblock In \emph{Proceedings of the IEEE/CVF conference on computer vision and pattern recognition}, 6458--6467.

\bibitem[{Li et~al.(2021{\natexlab{b}})Li, Wang, Liu, and Lin}]{AD1}
Li, T.; Wang, Z.; Liu, S.; and Lin, W.-Y. 2021{\natexlab{b}}.
\newblock Deep Unsupervised Anomaly Detection.
\newblock In \emph{Proceedings of the IEEE/CVF Winter Conference on Applications of Computer Vision (WACV)}, 3636--3645.

\bibitem[{Lin et~al.(2017)Lin, Doll{\'a}r, Girshick, He, Hariharan, and Belongie}]{lin2017feature}
Lin, T.-Y.; Doll{\'a}r, P.; Girshick, R.; He, K.; Hariharan, B.; and Belongie, S. 2017.
\newblock Feature pyramid networks for object detection.
\newblock In \emph{Proceedings of the IEEE conference on computer vision and pattern recognition}, 2117--2125.

\bibitem[{Liu et~al.(2021)Liu, Lin, Cao, Hu, Wei, Zhang, Lin, and Guo}]{liu2021swin}
Liu, Z.; Lin, Y.; Cao, Y.; Hu, H.; Wei, Y.; Zhang, Z.; Lin, S.; and Guo, B. 2021.
\newblock Swin transformer: Hierarchical vision transformer using shifted windows.
\newblock In \emph{Proceedings of the IEEE/CVF international conference on computer vision}, 10012--10022.

\bibitem[{Loshchilov and Hutter(2017)}]{adamw}
Loshchilov, I.; and Hutter, F. 2017.
\newblock Decoupled weight decay regularization.
\newblock \emph{arXiv preprint arXiv:1711.05101}.

\bibitem[{Nayef et~al.(2017)Nayef, Yin, Bizid, Choi, Feng, Karatzas, Luo, Pal, Rigaud, Chazalon et~al.}]{ic17mlt}
Nayef, N.; Yin, F.; Bizid, I.; Choi, H.; Feng, Y.; Karatzas, D.; Luo, Z.; Pal, U.; Rigaud, C.; Chazalon, J.; et~al. 2017.
\newblock Icdar2017 robust reading challenge on multi-lingual scene text detection and script identification-rrc-mlt.
\newblock In \emph{2017 14th IAPR international conference on document analysis and recognition (ICDAR)}, volume~1, 1454--1459. IEEE.

\bibitem[{Ojha, Li, and Lee(2023)}]{UniFD}
Ojha, U.; Li, Y.; and Lee, Y.~J. 2023.
\newblock Towards Universal Fake Image Detectors That Generalize Across Generative Models.
\newblock In \emph{Proceedings of the IEEE/CVF Conference on Computer Vision and Pattern Recognition (CVPR)}, 24480--24489.

\bibitem[{Pang et~al.(2021)Pang, Shen, Cao, and Hengel}]{AD2}
Pang, G.; Shen, C.; Cao, L.; and Hengel, A. V.~D. 2021.
\newblock Deep learning for anomaly detection: A review.
\newblock \emph{ACM computing surveys (CSUR)}, 54(2): 1--38.

\bibitem[{Peng et~al.(2023{\natexlab{a}})Peng, Liu, Liu, and Jin}]{peng2023viteraser}
Peng, D.; Liu, C.; Liu, Y.; and Jin, L. 2023{\natexlab{a}}.
\newblock ViTEraser: Harnessing the Power of Vision Transformers for Scene Text Removal with SegMIM Pretraining.
\newblock \emph{arXiv preprint arXiv:2306.12106}.

\bibitem[{Peng et~al.(2023{\natexlab{b}})Peng, Yang, Zhang, Liu, Shi, Ding, Guo, and Jin}]{peng2023upocr}
Peng, D.; Yang, Z.; Zhang, J.; Liu, C.; Shi, Y.; Ding, K.; Guo, F.; and Jin, L. 2023{\natexlab{b}}.
\newblock UPOCR: Towards Unified Pixel-Level OCR Interface.
\newblock arXiv:2312.02694.

\bibitem[{Perez-Cabo et~al.(2019)Perez-Cabo, Jimenez-Cabello, Costa-Pazo, and Lopez-Sastre}]{FASAD}
Perez-Cabo, D.; Jimenez-Cabello, D.; Costa-Pazo, A.; and Lopez-Sastre, R.~J. 2019.
\newblock Deep Anomaly Detection for Generalized Face Anti-Spoofing.
\newblock In \emph{Proceedings of the IEEE/CVF Conference on Computer Vision and Pattern Recognition (CVPR) Workshops}.

\bibitem[{Qu et~al.(2023{\natexlab{a}})Qu, Liu, Liu, Chen, Peng, Guo, and Jin}]{CVPR2023DocTamper}
Qu, C.; Liu, C.; Liu, Y.; Chen, X.; Peng, D.; Guo, F.; and Jin, L. 2023{\natexlab{a}}.
\newblock Towards robust tampered text detection in document image: new dataset and new solution.
\newblock In \emph{Proceedings of the IEEE/CVF Conference on Computer Vision and Pattern Recognition}, 5937--5946.

\bibitem[{Qu et~al.(2024{\natexlab{a}})Qu, Zhong, Guo, and Jin}]{qu2024omni}
Qu, C.; Zhong, Y.; Guo, F.; and Jin, L. 2024{\natexlab{a}}.
\newblock Omni-IML: Towards Unified Image Manipulation Localization.
\newblock \emph{arXiv preprint arXiv:2411.14823}.

\bibitem[{Qu et~al.(2024{\natexlab{b}})Qu, Zhong, Liu, Xu, Peng, Guo, and Jin}]{qu2024towards}
Qu, C.; Zhong, Y.; Liu, C.; Xu, G.; Peng, D.; Guo, F.; and Jin, L. 2024{\natexlab{b}}.
\newblock Towards Modern Image Manipulation Localization: A Large-Scale Dataset and Novel Methods.
\newblock In \emph{Proceedings of the IEEE/CVF Conference on Computer Vision and Pattern Recognition}, 10781--10790.

\bibitem[{Qu et~al.(2023{\natexlab{b}})Qu, Tan, Xie, Xu, Wang, and Zhang}]{AAAI_MOSTEL}
Qu, Y.; Tan, Q.; Xie, H.; Xu, J.; Wang, Y.; and Zhang, Y. 2023{\natexlab{b}}.
\newblock Exploring stroke-level modifications for scene text editing.
\newblock In \emph{Proceedings of the AAAI Conference on Artificial Intelligence}, volume~37, 2119--2127.

\bibitem[{Radford et~al.(2021)Radford, Kim, Hallacy, Ramesh, Goh, Agarwal, Sastry, Askell, Mishkin, Clark et~al.}]{clip}
Radford, A.; Kim, J.~W.; Hallacy, C.; Ramesh, A.; Goh, G.; Agarwal, S.; Sastry, G.; Askell, A.; Mishkin, P.; Clark, J.; et~al. 2021.
\newblock Learning transferable visual models from natural language supervision.
\newblock In \emph{International conference on machine learning}, 8748--8763. PMLR.

\bibitem[{Ren et~al.(2015)Ren, He, Girshick, and Sun}]{ren2015fasterrcnn}
Ren, S.; He, K.; Girshick, R.; and Sun, J. 2015.
\newblock Faster r-cnn: Towards real-time object detection with region proposal networks.
\newblock \emph{Advances in neural information processing systems}, 28.

\bibitem[{Roy et~al.(2020)Roy, Bhattacharya, Ghosh, and Pal}]{CVPR_STEFANN}
Roy, P.; Bhattacharya, S.; Ghosh, S.; and Pal, U. 2020.
\newblock STEFANN: Scene Text Editor Using Font Adaptive Neural Network.
\newblock In \emph{Proceedings of the IEEE/CVF Conference on Computer Vision and Pattern Recognition (CVPR)}.

\bibitem[{Shimoda et~al.(2021)Shimoda, Haraguchi, Uchida, and Yamaguchi}]{Shimoda_2021_ICCV}
Shimoda, W.; Haraguchi, D.; Uchida, S.; and Yamaguchi, K. 2021.
\newblock De-Rendering Stylized Texts.
\newblock In \emph{Proceedings of the IEEE/CVF International Conference on Computer Vision (ICCV)}, 1076--1085.

\bibitem[{Singh et~al.(2021)Singh, Pang, Toh, Huang, Galuba, and Hassner}]{singh2021textocr}
Singh, A.; Pang, G.; Toh, M.; Huang, J.; Galuba, W.; and Hassner, T. 2021.
\newblock Textocr: Towards large-scale end-to-end reasoning for arbitrary-shaped scene text.
\newblock In \emph{Proceedings of the IEEE/CVF conference on computer vision and pattern recognition}, 8802--8812.

\bibitem[{Sun et~al.(2019)Sun, Ni, Chng, Liu, Luo, Ng, Han, Ding, Liu, Karatzas et~al.}]{lsvt}
Sun, Y.; Ni, Z.; Chng, C.-K.; Liu, Y.; Luo, C.; Ng, C.~C.; Han, J.; Ding, E.; Liu, J.; Karatzas, D.; et~al. 2019.
\newblock ICDAR 2019 competition on large-scale street view text with partial labeling-RRC-LSVT.
\newblock In \emph{2019 International Conference on Document Analysis and Recognition (ICDAR)}, 1557--1562. IEEE.

\bibitem[{Sun et~al.(2023)Sun, Fang, Zhao, Wang, and Cao}]{sun2023rethinking}
Sun, Z.; Fang, H.; Zhao, X.; Wang, D.; and Cao, J. 2023.
\newblock Rethinking Image Editing Detection in the Era of Generative AI Revolution.
\newblock arXiv:2311.17953.

\bibitem[{Tan et~al.(2023)Tan, Zhao, Wei, Gu, and Wei}]{AIGCD1}
Tan, C.; Zhao, Y.; Wei, S.; Gu, G.; and Wei, Y. 2023.
\newblock Learning on Gradients: Generalized Artifacts Representation for GAN-Generated Images Detection.
\newblock In \emph{Proceedings of the IEEE/CVF Conference on Computer Vision and Pattern Recognition}, 12105--12114.

\bibitem[{Tuo et~al.(2023)Tuo, Xiang, He, Geng, and Xie}]{Anytext}
Tuo, Y.; Xiang, W.; He, J.-Y.; Geng, Y.; and Xie, X. 2023.
\newblock AnyText: Multilingual Visual Text Generation And Editing.
\newblock \emph{arXiv preprint arXiv:2311.03054}.

\bibitem[{Wang et~al.(2020{\natexlab{a}})Wang, Sun, Cheng, Jiang, Deng, Zhao, Liu, Mu, Tan, Wang et~al.}]{wang2020hrnetv2}
Wang, J.; Sun, K.; Cheng, T.; Jiang, B.; Deng, C.; Zhao, Y.; Liu, D.; Mu, Y.; Tan, M.; Wang, X.; et~al. 2020{\natexlab{a}}.
\newblock Deep high-resolution representation learning for visual recognition.
\newblock \emph{IEEE transactions on pattern analysis and machine intelligence}, 43(10): 3349--3364.

\bibitem[{Wang et~al.(2019{\natexlab{a}})Wang, Xie, Li, Hou, Lu, Yu, and Shao}]{wang2019psenet}
Wang, W.; Xie, E.; Li, X.; Hou, W.; Lu, T.; Yu, G.; and Shao, S. 2019{\natexlab{a}}.
\newblock Shape robust text detection with progressive scale expansion network.
\newblock In \emph{Proceedings of the IEEE/CVF conference on computer vision and pattern recognition}, 9336--9345.

\bibitem[{Wang et~al.(2019{\natexlab{b}})Wang, Jiang, Luo, Liu, Choi, and Kim}]{wang2019atrr}
Wang, X.; Jiang, Y.; Luo, Z.; Liu, C.-L.; Choi, H.; and Kim, S. 2019{\natexlab{b}}.
\newblock Arbitrary shape scene text detection with adaptive text region representation.
\newblock In \emph{Proceedings of the IEEE/CVF conference on computer vision and pattern recognition}, 6449--6458.

\bibitem[{Wang et~al.(2022)Wang, Xie, Xing, Wang, Zhu, and Zhang}]{wang2022tic13}
Wang, Y.; Xie, H.; Xing, M.; Wang, J.; Zhu, S.; and Zhang, Y. 2022.
\newblock Detecting tampered scene text in the wild.
\newblock In \emph{European Conference on Computer Vision}, 215--232. Springer.

\bibitem[{Wang et~al.(2020{\natexlab{b}})Wang, Xie, Zha, Xing, Fu, and Zhang}]{wang2020contournet}
Wang, Y.; Xie, H.; Zha, Z.-J.; Xing, M.; Fu, Z.; and Zhang, Y. 2020{\natexlab{b}}.
\newblock Contournet: Taking a further step toward accurate arbitrary-shaped scene text detection.
\newblock In \emph{proceedings of the IEEE/CVF conference on computer vision and pattern recognition}, 11753--11762.

\bibitem[{Wang et~al.(2023)Wang, Bao, Zhou, Wang, and Li}]{DFD2}
Wang, Z.; Bao, J.; Zhou, W.; Wang, W.; and Li, H. 2023.
\newblock Altfreezing for more general video face forgery detection.
\newblock In \emph{Proceedings of the IEEE/CVF Conference on Computer Vision and Pattern Recognition}, 4129--4138.

\bibitem[{Wu et~al.(2019)Wu, Zhang, Liu, Han, Liu, Ding, and Bai}]{ACMMM_SRNet}
Wu, L.; Zhang, C.; Liu, J.; Han, J.; Liu, J.; Ding, E.; and Bai, X. 2019.
\newblock Editing Text in the Wild.
\newblock In \emph{Proceedings of the 27th ACM International Conference on Multimedia}, MM '19, 1500–1508. New York, NY, USA: Association for Computing Machinery.
\newblock ISBN 9781450368896.

\bibitem[{Xie et~al.(2021)Xie, Wang, Yu, Anandkumar, Alvarez, and Luo}]{xie2021segformer}
Xie, E.; Wang, W.; Yu, Z.; Anandkumar, A.; Alvarez, J.~M.; and Luo, P. 2021.
\newblock SegFormer: Simple and efficient design for semantic segmentation with transformers.
\newblock \emph{Advances in neural information processing systems}, 34: 12077--12090.

\bibitem[{Zhang et~al.(2019)Zhang, Zhou, Jiang, Song, Li, Zhou, Wang, Wang, Liao, Yang et~al.}]{rects}
Zhang, R.; Zhou, Y.; Jiang, Q.; Song, Q.; Li, N.; Zhou, K.; Wang, L.; Wang, D.; Liao, M.; Yang, M.; et~al. 2019.
\newblock Icdar 2019 robust reading challenge on reading chinese text on signboard.
\newblock In \emph{2019 international conference on document analysis and recognition (ICDAR)}, 1577--1581. IEEE.

\bibitem[{Zhao and Lian(2023)}]{Udifftext}
Zhao, Y.; and Lian, Z. 2023.
\newblock UDiffText: A Unified Framework for High-quality Text Synthesis in Arbitrary Images via Character-aware Diffusion Models.
\newblock \emph{arXiv preprint arXiv:2312.04884}.

\bibitem[{Zhou et~al.(2023)Zhou, Zhang, Yao, Lu, Yi, Ding, and Ma}]{FAS2}
Zhou, Q.; Zhang, K.-Y.; Yao, T.; Lu, X.; Yi, R.; Ding, S.; and Ma, L. 2023.
\newblock Instance-aware domain generalization for face anti-spoofing.
\newblock In \emph{Proceedings of the IEEE/CVF Conference on Computer Vision and Pattern Recognition}, 20453--20463.

\bibitem[{Zhou et~al.(2017)Zhou, Yao, Wen, Wang, Zhou, He, and Liang}]{zhou2017east}
Zhou, X.; Yao, C.; Wen, H.; Wang, Y.; Zhou, S.; He, W.; and Liang, J. 2017.
\newblock East: an efficient and accurate scene text detector.
\newblock In \emph{Proceedings of the IEEE conference on Computer Vision and Pattern Recognition}, 5551--5560.

\end{thebibliography}
